\pdfoutput=1

\documentclass[11pt]{article}
\usepackage{comment}
\usepackage{verbatim}
\usepackage{bbm}
\usepackage{tikz}
\usepackage{xtab}
\usepackage{multirow, makecell}
\usepackage{amsmath}
\usepackage{graphicx} 
\usepackage{wrapfig}
\usepackage{subcaption}
\usepackage{xcolor,colortbl}
\usepackage{algorithm}
\usepackage{array}
\usepackage{microtype}
\usepackage{graphicx}
\usepackage{longtable}
\usepackage{multirow}
\usepackage{booktabs} 
\usepackage{xcolor,colortbl}

\usepackage{color-edits}
\usepackage{comment}
\addauthor{kb}{purple}

\usepackage{hyperref}
\definecolor{highlightblue}{RGB}{173,216,230} 
\definecolor{highlightgreen}{RGB}{144,238,144} 
\definecolor{mediumblue}{RGB}{100,149,237} 
\definecolor{mediumgreen}{RGB}{60,179,113} 
\definecolor{mediumred}{RGB}{240,128,128} 

\usepackage{algorithmic}

\definecolor{Gray}{gray}{0.9}
\newcolumntype{a}{>{\columncolor{Gray}}c}

\usepackage[utf8]{inputenc} 
\usepackage[T1]{fontenc}    
\usepackage{hyperref}       
\usepackage{url}            
\usepackage{booktabs}       
\usepackage{amsfonts}       
\usepackage{nicefrac}       
\usepackage{microtype}      

\usepackage[preprint]{acl}

\usepackage{times}
\usepackage{latexsym}

\usepackage[T1]{fontenc}

\usepackage[utf8]{inputenc}

\usepackage{microtype}

\usepackage{inconsolata}

\usepackage{graphicx}

\usepackage{amsmath}
\usepackage{amssymb}
\usepackage{mathtools}
\usepackage{amsthm}

\usepackage[capitalize,noabbrev]{cleveref}

\theoremstyle{plain}
\newtheorem{theorem}{Theorem}[section]

\theoremstyle{definition}

\theoremstyle{remark}
\newtheorem{remark}[theorem]{Remark}

\usepackage[textsize=tiny]{todonotes}
\title{CLIPer: Tailoring Diverse User Preference via Classifier-Guided Inference-Time Personalization}

\author{Jinyan Su, Jinpeng Zhou,
  Claire Cardie, Wen Sun\\
Cornell University\\
  \texttt{\{js3673, jpzhou, ctc9, ws455\}@cornell.edu}
  }

\begin{document}
\maketitle

\begin{abstract}
Personalized LLMs can significantly enhance user experiences by tailoring responses to preferences such as helpfulness, conciseness, and humor. However, fine-tuning models to address all possible combinations of user preferences is computationally expensive and impractical. In this paper, we introduce \textbf{CLIPer}(\textbf{Cl}assifier-guided \textbf{I}nference-time \textbf{Per}sonalization), a lightweight personalization approach that leverages a classifier model to steer LLM generation dynamically to different user preferences at inference time. Our method eliminates the need for extensive fine-tuning, inducing negligible additional computational overhead while enabling more controllable and nuanced personalization across single and multi-dimensional preferences. Comprehensive empirical analyses demonstrate the scalability and effectiveness of our approach in delivering personalized language generation.
\end{abstract}

\section{Introduction}
 In recent years, LLMs have emerged as powerful tools, driving advancements across various applications. Their potential for delivering personalized services to meet diverse user needs has spurred significant interest in developing personalized LLMs, with the ultimate goal of leveraging them as personal assistants. While methods like Reinforcement Learning from Human Feedback (RLHF) have aligned LLMs with general human preferences, they fall short in addressing the vast diversity of individual user preferences. This generalization can negatively impact the user experience, particularly for minority groups, and limit the accessibility of LLMs. For instance, an LLM designed for children should be amicable and easy to understand, whereas one for IT professionals should employ technical and formal language.

\begin{figure}[h]
    \centering
\includegraphics[width=\linewidth]{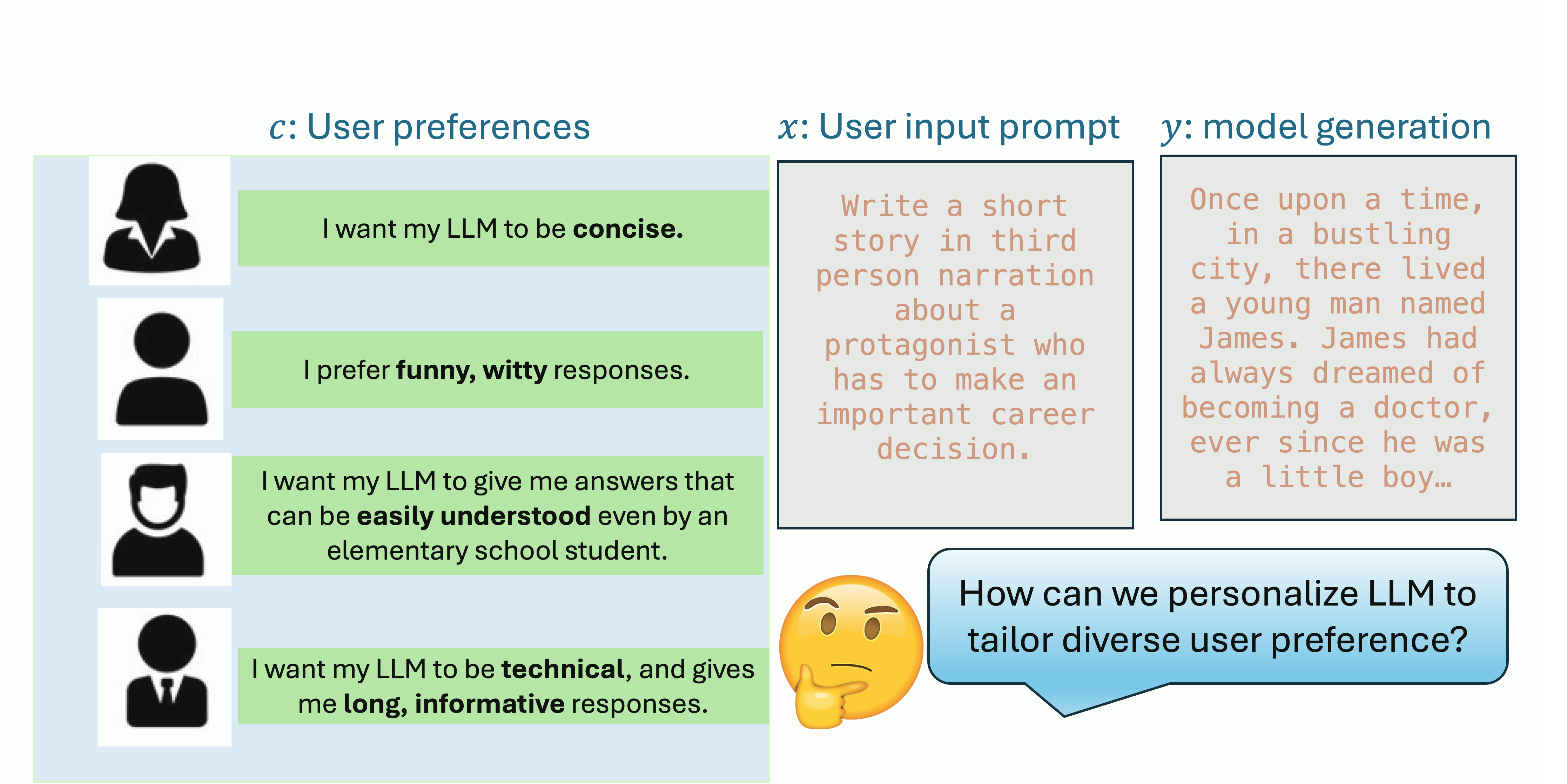}
    \caption{Motivation: Different users have different preferences.}
    \label{fig: personalziatoin}
\end{figure}

\begin{table}[t!]
\centering
\scriptsize
\begin{tabular}{cccc}
\toprule
Methods& Training Cost& Inference Cost\\
\toprule
Naive Fine-tuning&$O(2^d)$ & $O(1)$\\
\hline
P-Soup \citep{jang2023personalized}
&$O(d)$& $O(1)$\\
\hline
MoPE \citep{zhou2024orchestrating}
& $O(d)$& $O(d)$ \\
\hline
CLIPer(Ours)
& $O(1)$&$O(1)$\\
\bottomrule
\end{tabular}
\caption{Compare different personalization approaches. Let $d$ be number of preferences user can choose from, such as \{\texttt{conciseness, funny, informal,...}\}, for each preference dimension, user can choose to have it or not, leading to $2^d$ total options in total.  Naive fine-tuning requires to train $2^d$ models in total. Meanwhile, both \cite{jang2023personalized}  and \cite{zhou2024orchestrating} fine-tunes models for each dimension, thus, training $d$ models in total. Moreover, \cite{zhou2024orchestrating} requires all the $d$ models during inference, resulting in $O(d)$ inference time cost. Our method doesn't require training any model for the preference dimensions.}
\label{tab: compared to previsou method}
\end{table}
Personalization in LLMs (Figure \ref{fig: personalziatoin}) offers the potential to significantly enhance user experiences by tailoring the model’s behavior to align with diverse individual or group preferences. A practical way to approach this is by clustering user preferences along various dimensions, such as helpfulness, conciseness, or humor. Users can then define their ideal LLM personality by choosing among these dimensions, allowing the model to adapt dynamically to their needs.

The simplest approach for personalization is through direct prompting, i.e., we include the user preference into the prompt. However, preference prompting can struggle especially when preference cannot be explicitly articulated in natural language, but can only be inferred from user preference data. In such cases, the most straightforward approach is to fine-tune a model on data reflecting each user's preferences. However, fine-tuning a model for each preference type already demands significant computational and memory resources. 
The challenge intensifies when user preferences span multiple dimensions, as the number of possible combinations grows exponentially. For example, as illustrated in Figure \ref{fig: personalziatoin}, some user might want their LLM to be concise, some might prefer their models to be funny, while some might want their models to have combinations of dimensions such as to be both concise and funny. 
Prior work \cite{jang2023personalized, zhou2024orchestrating} acknowledges the necessity of fine-tuning models on single dimension, and mainly focus on reducing the complexity when users have combined preferences along multiple dimensions, reducing the training cost from exponential to linear in terms of dimension $d$. See the comparison in Table \ref{tab: compared to previsou method}.

\begin{figure*}[h]
    \centering
\includegraphics[width=\linewidth]{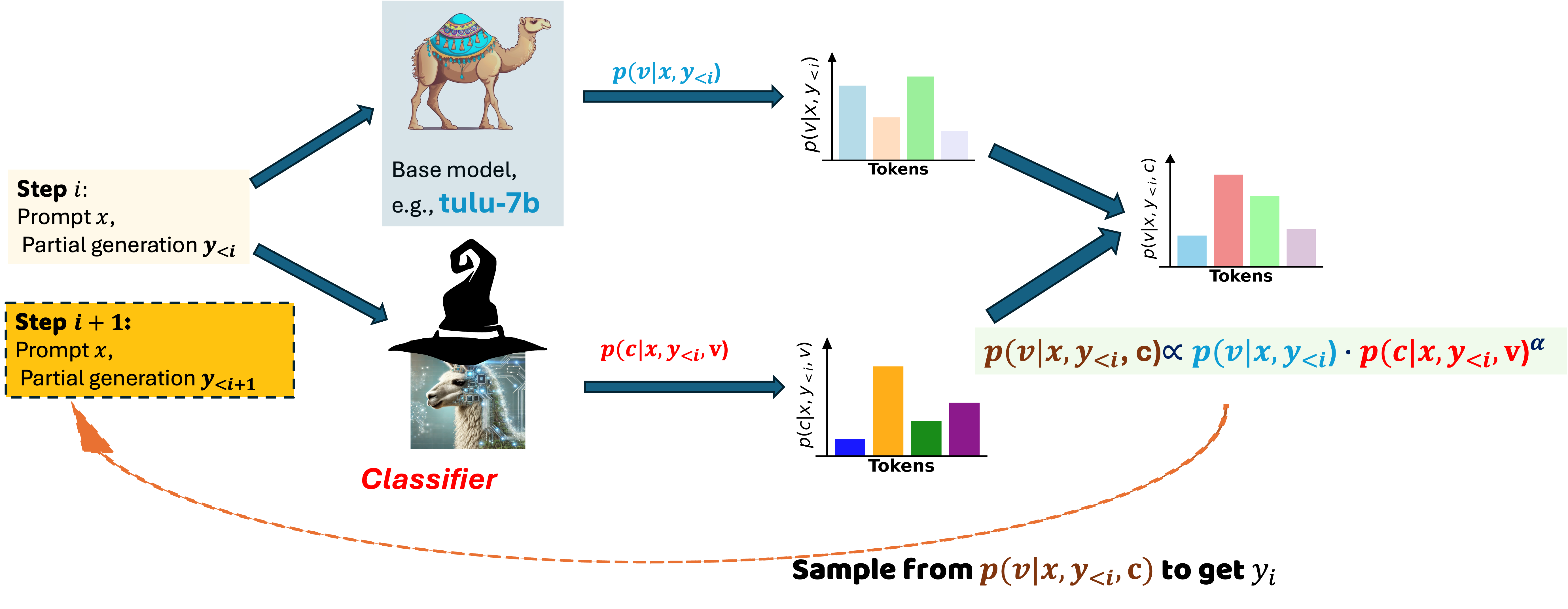}
    \caption{Overview of CLIPer: at each step $i$, a lightweight classifier model outputs the probability of the preferences classes given prompt $x$, partial generation in previous steps $y_{<i}$ and potential next token. The detail can be found in Section \ref{Sec 3.2}.  }
    \label{fig: modification}
\end{figure*}

However, training $d$ separate models for each preference dimension can be impractical for for large LLMs. 
To make personalization more feasible, we need a lightweight personalization approach without fine-tuning LLM for every preference dimension. In this paper, 
we propose \textbf{CLIPer}, a \textbf{Cl}assifier-guided \textbf{I}nference-time \textbf{Per}sonalization approach that eliminates the need for training and maintaining multiple fine-tuned models during both training and inference. Our classifier is trained to predict a distribution over all preference dimensions given the past text.  The distribution models how likely the given text  can be classified as each preference dimension (e.g., what is the probability that the given text can be classified as funny). Our classifier predicts at the token level, and its output can be used to dynamically adjust the base model's logits at token level during the inference time. Our method incurs negligible computational overhead compared to fine-tuning multiple large LLMs while effectively steers the LLM toward personalized behaviors.

To summarize, our contributions are:\\
(1)  We propose a \textbf{Cl}assifier-guided \textbf{I}nference-time \textbf{Per}sonalization approach (\textbf{CLIPer}) that trains a small classifier which dynamically adjusts the logits of the LLM during inference time, enabling personalization without fine-tuning the LLM. This approach maintains a constant training and inference cost with respect to preference dimensions, making it both scalable and practical for real-world applications. \\
(2) Our method supports both single-dimension and multi-dimension preference settings.\\
(3) We introduce a novel classifier  model architecture that reduces computational overhead by jointly predicting the conditional distribution over the preference dimensions conditioned on past tokens (prefix) at the same time. 
This architecture enables our approach to perform only one forward pass on both the large LLM and the small classifier during the inference time. \\
(4) Through comprehensive empirical analyses, we demonstrate the effectiveness of our approach in achieving personalized and controllable text generation without the need for extensive fine-tuning or additional computational burden.

\section{Related Work}
\paragraph{Personalized LLM.}  P-RLHF \cite{li2024personalized} introduces a lightweight user model to capture individual preferences and jointly learns both the user model and the personalized LLM from human feedback. \citet{poddar2024personalizing} infer a user-specific latent representation, using it to condition both the reward model and policy—without requiring additional user-specific data. \citet{zhuanghydra} propose a system that trains a reranker to prioritize useful information and an adapter to align outputs with user preferences. Similarly \citet{zollopersonalllm} simulate a large and diverse user base by leveraging pre-trained reward models to generate user-specific signals.
\citet{jang2023personalized} proposed Personalized Soups (P-Soup), reducing the training cost to linear complexity with respect to the number of dimensions $d$. Subsequent work by \citet{zhou2024orchestrating} use inference time personalization, 
merging the outputs of dimension-wise expert models. However, both \citet{jang2023personalized} and \citet{zhou2024orchestrating} require training an expert model for each dimension, resulting in $d$ additional training computations for $d$ dimensions. Furthermore, \citet{zhou2024orchestrating} employs $d$ expert models simultaneously during inference, leading to $O(d)$ computational complexity for inference, as summarized in Table \ref{tab: compared to previsou method}. 

\paragraph{LLM Alignment} \citet{dathathriplug} introduce Plug and Play Language Models (PPLM), which enable controllable generation by combining a pretrained LLM with lightweight attribute classifiers—without retraining the base model. \citet{li2023inference} propose inference-time interventions to elicit more truthful responses. \citet{khanovargs} guide generation using a learned reward function, while \citet{chakraborty2024transfer} estimate the optimal value function for a target reward implicitly. \citet{mudgal2024controlled} train a separate prefix scorer to approximate the reward function. \citet{chen2024pal} propose PAL, a general framework for pluralistic alignment that explicitly incorporates diverse human preferences.  Several other works also explore alignment through a multi-objective reinforcement learning lens \citet{shi2024decoding, rame2023rewarded, yangmetaaligner} drawing on foundations from (multi-objective) RL \cite{sener2018multi}. 
\section{Method}
\subsection{Problem Setting}
\paragraph{Preferences Dimensions}
Given a set of $d$ preferences, for example, $\mathcal{P}=$\{\texttt{Concise, Funny, Rigorous...}\}, a user can choose among these $d$ preference dimensions. Let $c$ be a single user preference dimension, for instance, $c=$\texttt{Concise} or $c=$\texttt{Funny}. Let $\textbf{c}$ be a set of user preferences, for example, $\textbf{c}=$\{\texttt{Concise}, \texttt{Funny}\}, indicating that user wants their model to be both concise and funny. The total number of different $\textbf{c}$ is $2^d$.

\paragraph{Independence Assumption} 
Denote $\mathcal{P} = \{c_1, \cdots, c_d\}$, and $P( c_i | x, y )$ for $i = 1,2,\dots, d$ as the probability of the prompt and the partial sentence $y$ being classified as preference $c_i$.  
We assume user preferences  in $\textbf{c}$ are independent, i.e., for any preference set $\textbf{c}$, we assume $P( \textbf{c} | x, y ) = \prod_{c\in \textbf{c}} P(c | x, y)$. 
This independence assumption 
enables our approach to scale efficiently with $d$ when $c$ spans multiple dimensions--- Instead of training a classifier for all possible $\textbf{c}$ among $2^d$ choices, we only train $d$-wise classifier, each corresponding to a single preference dimension.


\subsection{Personalized Text Generation}\label{Sec 3.2}
Given a user preference $c$ and prompt $x$ (such as shown in Figure \ref{fig: personalziatoin}), the text generation from personalized model can be formulated  as a conditional text probability of generation $y$ given $x$ and $c$: 
\begin{equation}
p(y|x,c)=\Pi_{i=1}^n p(y_i|y_{<i}; x,c),
\end{equation}
where $y_i$ denotes the $i$-th token in the generated text and $y_{<i}$ represents the tokens generated up to step $i-1$.

While training a separate model for each possible preference 
$c$ would theoretically enable perfect personalization, it is computationally infeasible due to the exponential growth of required models as the number of preference dimensions $d$ increases. Instead, we focus on inference-time personalization, where a general base model $p(y|x)$ is dynamically adjusted to approximate the personalized model $p(y|x,c)$.

To achieve this, we leverage Bayesian inference by introducing an auxiliary classifier  model $p(c|y, x)$ to steer the base model:
\begin{equation}\label{eq: 1}
    p(y| x, c)\propto p(y|x) \cdot p(c|y,  x)^{\alpha},
\end{equation}
where  $\alpha$ is a hyperparameter controlling the influence of the classifier model. When $\alpha=0$, the formulation reduces to the base model without personalization.

In the case of autoregressive language models, where $p(y|x,c)=\Pi_{i=1}^n p(y_i|y_{<i}; x,c)$, the conditional probability for each token is:
\begin{equation}\label{eq: 2}
    p(y_i|y_{<i}; x, c)\propto p(y_i|y_{<i};x) \cdot p(c|y_{\leq i}, x)^{\alpha}.
\end{equation}
Thus, the overall personalized text generation probability becomes:
\begin{equation*}
    p(y|x, c)\propto \Pi_{i=1}^n p(y_i|y_{<i};x) p(c|y_{\leq i}, x)^{\alpha}
\end{equation*} 
Here, $p(c|y_i, y_{<i}, x)$ represents the likelihood of preference $c$ given prompt $x$ and the partial generation $y_{\leq i}$, which are given by the classifier model. 
\paragraph{Multiple Preferences} It is intuitive to generalize the above formulation to multiple preferences, i.e., $\textbf{c}=\{c_1, \cdots, c_d\}$, and $|\textbf{c}|=k$. Using the independence assumption, we have 
\begin{equation}\label{eq: 3}
\small
p(y_i|y_{<i}; x, \textbf{c})\propto p(y_i|y_{<i};x)\Pi_{c_k\in \mathbf{c}}p(c_k|y_{\leq i};x)^{\alpha_k}
\end{equation}
where $\alpha = (\alpha_1, \cdots, \alpha_d)$ are hyperparameters controlling the influence of the each preference.

In inference time, to sample the next token, we need to construct the distribution over the next token (i.e., $p(y | y_{<i}; x, c)$ for all $y\in \mathcal{V}$). This would require us to evaluate the classifier model for all $y\in \mathcal{V}$, i.e., we need to compute $p(c | y_{<i}, y, x)$ for all $y\in \mathcal{V}$. Naively implementing this would require us to perform $|\mathcal{V}|$ forward passes on the classifier model \citep{sitdikov2022classifiers}.  This can significantly slow down the inference procedure. 


\subsection{Efficient classifier Modeling}\label{sec: guidance modeling}
To address the above efficiency challenge, we train a classifier model capable of producing classification probabilities for all tokens in the vocabulary at each step. Specifically, during the generation, the classifier model takes the prompt $x$ and partial generation $y_{<i}$ as inputs, and outputs a matrix $M\in \mathbb{R}^{|\mathcal{V}|\times d}$ where $M_{v,c_j}$ models the probability of having label $c_j$ given prompt $x$, the partial generation $(y_{<i})$, and the next token $v\in \mathcal{V}$, i.e., $p(c_j  | y_i = v  , y_{<i}, x)$. 
\begin{figure*}[h]
    \centering
\includegraphics[width=\linewidth]{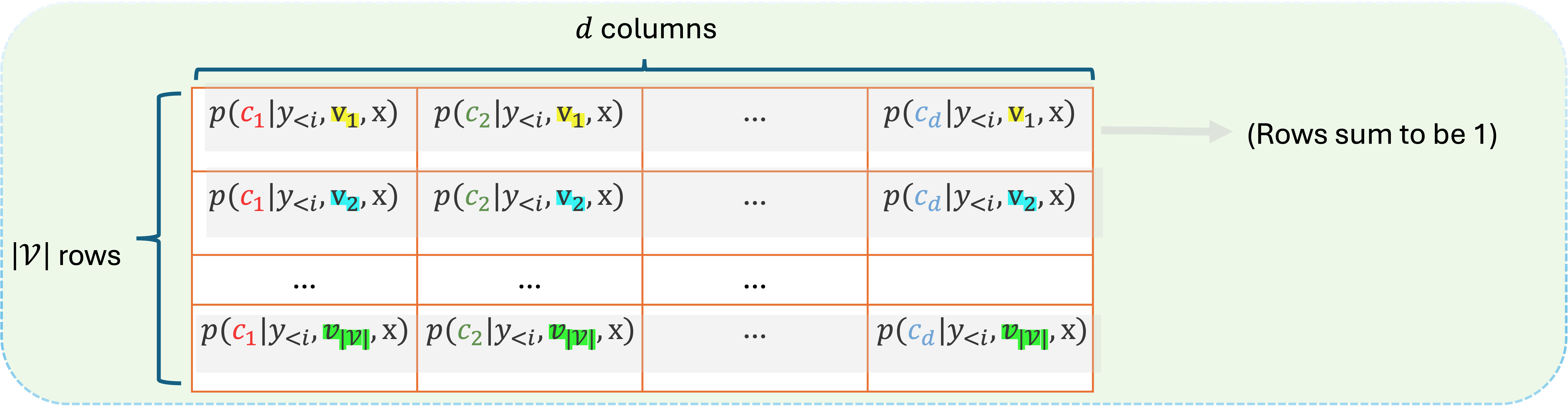}
    \caption{Details of the output matrix $M$. Given $y_{<i}$, each row of $M$ provides the preference probabilities for a specific token in the vocabulary, which sums up to be 1.}
    \label{fig: classifier}
\end{figure*}
The $k$-th row of $M$ corresponds to the preference probabilities $p(c|y_{< i}, v_k, x)$, where $v_{k}\in \mathcal{V}$ is the $k$-th token in the vocabulary. The rows are normalized such that the probabilities sum to 1. 
This design allows the classifier model to efficiently compute preference probabilities for all possible tokens in a single step, avoiding the need for performing multiple forward passes of the model. 
An illustrative representation of the matrix $M$ is depicted in Figure \ref{fig: classifier}.

\paragraph{Training Loss} 
Given $k$ datasets where each dataset $\mathcal{D}_i$ contains generations that align well with the given preference dimension $c_i$. 
The classifier model is trained using a combined dataset $D=\cup_iD_i$. Let the combined $D= \{(x^{(j)}, y^{(j)}, c^{(j)})\}$, where each sample consists of a prompt $x^{(j)}$, a generated text $y^{(j)}$, and a preference $c^{(j)}\in \mathcal{P}$.
The training objective is to minimize the following token-level multi-class classification loss:
\begin{equation}
\small
\mathcal{L} = -\frac{1}{N}\sum_{j=1}^N \left(\frac{1}{n^{(j)}}\sum_{i=1}^{n^{(j)}} \log p(c^{(j)}|y^{(j)}_{\leq i}, x^{(j)})\right),
\label{eq:loss}
\end{equation}
where $N$ is the total training samples, $n^{(j)}$ is the total number of tokens of $y^{(j)}$. 
We show the construction of the training loss through a concrete example in Figure \ref{fig: loss}. 

\subsection{Classifier Model Guided Generation.}
After training the classifier model, we utilize it to achieve personalized generation as described in Eq. (\ref{eq: 1}). At each generation step $i$, the base model and the classifier model both take the prompt $x$ and the current partial generation $y_{<i}$ as input. Then, produces the unconditional probabilities $p(v|x, y_{<i})$ for all tokens $v\in \mathcal{V}$, where $\mathcal{V}$ is the vocabulary. And the classifier model generates probabilities of user preference $c$, i.e.,  $p(c|x, y_{<i}, v)$, for all $v\in \mathcal{V}$. The next token $y_i$ is sampled from the personalized probability distribution $p(v|x,y_{<i}, c)$, which combines the outputs of the base model and the classifier model through Eq.(\ref{eq: 2}) (Or Eq.(\ref{eq: 3}) for multiple preferences). The process is repeated until the end of the generation 
An illustration of this classified guided process is provided in Figure \ref{fig: modification}.

\section{Experiments}
\subsection{Experimental Detail} 
\paragraph{Dataset for Classifier Model Training}
Following prior work \cite{zhou2024orchestrating, jang2023personalized}, we consider six preference dimensions across three categories: audience, content density, and tone (see Table \ref{tab: preference prompt for each dimension}). Consistent with previous studies \cite{jang2023personalized, zhou2024orchestrating}, we use GPT4-Alpaca \cite{peng2023instruction} as training data and generate $n=16$ samples per dimension using Tulu-7B as the base model, with prompts from \cite{jang2023personalized} (summarized in Table \ref{tab: preference prompt for each dimension}). The top-1 ranked sample, based on the reward model from \cite{jang2023personalized}, is selected as training data for each dimension. We then train a classifier using LLaMA-160M on this collected data.
Unlike \cite{jang2023personalized, zhou2024orchestrating}, we do not use reward model for training or inference; it is only employed during dataset creation to simulate human preferences.

Due to the scarcity of high-quality validation and test datasets, we tune hyperparameters $\alpha$ using data randomly sampled from Alpaca-Eval. For evaluation, we follow prior works \cite{jang2023personalized, zhou2024orchestrating} and use the Koala and UltraFeedback \cite{cui2024ultrafeedback} datasets.

\begin{table*}[t!]
\centering
\scriptsize
\begin{tabular}{cccc}
\toprule
Category& Symbol& Dimension& Prompt\\
\toprule
\multirow{2}{*}{
\textbf{Audience-specific}
}&P1A& Elementary Level & Generate a response that can be easily understood by an elementary school student.\\
&P2A &PhD Level& Generate a response that only a PhD Student in that specific field could understand.\\
\hline
\multirow{2}{*}{
\textbf{Content-density}
}& P2A& Concise& Generate a response that is concise and to the point without being verbose.\\
&P2B& Verbose&  Generate a response that is very informative without missing any background information.\\
\hline
\multirow{2}{*}{
\textbf{Tone}
}& P3A&Playful &Generate a response that is friendly, witty, funny, and humorous, like a close friend.\\
& P3B&Harsh&Generate a response in an unfriendly manner.\\
\bottomrule
\end{tabular}
\caption{Preference symbol, their corresponding dimension, as well as the prompt that is used for direct prompting. }
\label{tab: preference prompt for each dimension}
\end{table*}
\paragraph{Preference Dimension Setting}

In order to accommodate the experiments to an important baseline - direct prompting with the preferences, and also to be consistent with previous works, we use explicit preferences, where the 6 dimensions are, "\texttt{elementary level}", "\texttt{PhD level}", "\texttt{concise}", "\texttt{verbose}", "\texttt{playful}", "\texttt{harsh}", respectively. The details description of these dimensions can be found in Table \ref{tab: preference prompt for each dimension}.

\paragraph{Baselines}
We mainly compare our method to 3 baselines: 
\textit{Direct Prompting}, which directly prompt the LLM to generate the text according to user preferences. Noting from prior works \cite{jang2023personalized, zhou2024orchestrating}, Direct prompting is a rather strong baseline since tulu-7b is well instructed to generate personalized texts based on instructions alone. The prompt used for each dimension is illustrated in Table \ref{tab: preference prompt for each dimension}.
\textit{p-soup \& Direct Fine-tuning.} For single-dimension experiments, we compare with fine-tuned models. For multi-dimension experiments, we compare with p-soup \cite{jang2023personalized}, which merges the parameters of models fine-tuned on each dimensions. 
\textit{Vanilla Baseline.} Texts generated directly from the base model, without tailoring to any user preference.

\paragraph{Evaluation Metrics} Since our classifier model is trained on dataset created with reward models, we use the reward model as a judge for win-rate. Additionally, we employ GPT-4o-mini as an alternative judge to simulate an out-of-domain preference scenario—where preferences are learned from one user group (i.e., the reward model) but evaluated on another (i.e., GPT-4o-mini). This further underscores the importance of personalizing LLMs for different user groups. We used the same prompts from \cite{jang2023personalized, zhou2024orchestrating} for GPT-4o-mini when using it as a judge. 
\begin{table*}[h]
\centering
\begin{tabular}{c|cc|cc|c}
\toprule
Evaluator& \multicolumn{2}{c|}{\textbf{Reward Model}} & \multicolumn{2}{c|}{\textbf{GPT-4o-mini}} & \multirow{2}{*}{\textbf{Average}} \\\cline{1-5}
Dataset& Koala & UltraFeedback & Koala & UltraFeedback &  \\
\midrule
\textit{Direct Prompting} & 0.5367 & 0.5367 & 0.5500 & 0.4967 & \cellcolor{highlightblue}0.5300 \\
\textit{Vanilla Baseline} & 0.8800 & 0.7800 & 0.8433 & 0.8333 & \cellcolor{highlightblue}0.8342 \\
\textit{p-soup \& Direct Fine-tuning} & 0.7867 & 0.7133 & 0.3967 & 0.4233 & \cellcolor{highlightblue} 0.5800 \\
\bottomrule
\end{tabular}
\caption{Average win rate on Koala and UltraFeedback, evaluated by reward model and GPT-4o-mini respectively. (Preference for 1-dim).}
\label{tab:combined_results}
 \vspace{-0.5cm}
\end{table*}

\begin{table*}[ht]
\scriptsize
\centering
\begin{tabular}{c|cc|cc|c}
\toprule
Evaluator & \multicolumn{2}{c|}{\textbf{Reward Model}} & \multicolumn{2}{c|}{\textbf{GPT-4o-mini}} & \multirow{2}{*}{\textbf{Average}} \\\cline{1-5}
Dataset & Koala & UltraFeedback & Koala & UltraFeedback &  \\
\midrule
\textit{Direct Prompting} & 0.5275 & 0.4844 & 0.5031 & 0.5038 & \cellcolor{highlightblue}0.5047 \\
\textit{Vanilla Baseline} & 0.7775 & 0.7125 & 0.7106 & 0.6975 & \cellcolor{highlightblue}0.7245 \\
\textit{p-soup \& Direct Fine-tuning} & 0.7063 & 0.6550 & 0.4775 & 0.4537 & \cellcolor{highlightblue}0.5731 \\
\bottomrule
\end{tabular}
\caption{Average win rate on Koala and UltraFeedback, evaluated by reward model and GPT-4o-mini respectively. (Preference for 2-dim).}
\label{tab:2d_combined_results}
\end{table*}

\begin{table*}[h]
\scriptsize
\centering
\begin{tabular}{c|cc|cc|c}
\toprule
Evaluator & \multicolumn{2}{c|}{\textbf{Reward Model}} & \multicolumn{2}{c|}{\textbf{GPT-4o-mini}} & \multirow{2}{*}{\textbf{Average}} \\\cline{1-5}
Dataset & Koala & UltraFeedback & Koala & UltraFeedback &  \\
\midrule
\textit{Direct Prompting} & 0.5933 & 0.4600 & 0.5367 & 0.5367 & \cellcolor{highlightblue}0.5317 \\
\textit{Vanilla Baseline} & 0.9000 & 0.7900 & 0.8067 & 0.8100 & \cellcolor{highlightblue}0.8267 \\
\textit{p-soup \& Direct Fine-tuning} & 0.8533 & 0.7533 & 0.5467 & 0.5833 & \cellcolor{highlightblue}0.6842 \\
\bottomrule
\end{tabular}
\caption{Average win rate on Koala and UltraFeedback, evaluated by reward model and GPT-4o-mini respectively. (Preference for 3-dim).}
\label{tab:3d_combined_results}
 \vspace{-0.5cm}
\end{table*}

\paragraph{Hyperparameter} The primary hyperparameter we tune is $\alpha$, which controls the influence of the classifier model in modifying the base model’s logits. A lower 
$\alpha$ results in less personalization, allowing the base model’s original responses to dominate, while a higher 
$\alpha$ enforces stronger personalization. In practice, 
$\alpha$ should be chosen to balance the trade-off between preserving the base model’s original responses and incorporating personalized adjustments. For CLIPer, $\alpha$ is a tuple, depending on total number of dimensions in the combination $\textbf{c}$. For single dimension experiments, $\alpha$ is a scalar, so we use two randomly sampled subset of alpaca eval dataset, each of 100 data samples,  and conduct parameter search over $\alpha=\{0.05, 0.1, 0.3, 0.5, 0.8\}$. We end up having $\alpha=0.5$ to be the best parameter for one subset and $\alpha=0.8$ for another. We end up choosing $\alpha=0.8$ as the hyper-parameter for our main experiments by taking the average of the win rate of these two dataset, and select the hyper-parameter based on this averaged win rate. Note that, since the win rate is pair-wise, to simplify the hyper-parameter selection, we only compute the win-rate against direct-prompting, and use this win rate to choose hyper-parameter. Also, we use the same hyper-parameter for all the experiments, thus, our hyper-parameter selection is rather coarse. For 2-dimension and 3 dimension experiments, with the experience from single dimensional hyper-parameter selection, we narrow down our search space and select $\alpha_i$ only from $\{0.5, 0.8\}$. Note that for 2-dimensional experiments, there are two hyper-parameters $(\alpha_1, \alpha_2)$, each for one dimension, while for 3-dimensional experiments, there are three hyper-parameters $(\alpha_1, \alpha_2, \alpha_3)$, thus, we search over 8 total combination of hyper-parameters.

\section{Experimental Results}
\subsection{Main Results}
In this section, we experiment when users choose one, two and three combined preference out of the 6 dimensions $\mathcal{P} =$\{P1A, P1B, P2A, P2B, P3A, P3B\}. 
\paragraph{Single preference Dimension} For single preference dimension experiments, we only have 1 hyper-parameter to tune. In order to show the robustness of our method, we present the average performance over these 6 dimensions in Table \ref{tab:combined_results} without tuning $\alpha$ for different dimensions. Even without carefully tuning $\alpha$, we find that CLIPer still achieves a rather high average win rate compare to other baseline.
Table \ref{tab:combined_results} presents the average win rates (across six single preference dimensions) from both reward model and gpt-4o-mini evaluations, demonstrating the effectiveness of our proposed approach. When compared with vanilla baseline, our method achieves a high average win rate of 83.42\%, highlighting the necessity of personalizing LLMs. Compared to Direct Prompting baseline, CLIPer attains  an average win rate of 53\%, while achieving 58\% average win rate against Direct fine-tuning. When comparing against direct prompting, there is minimal discrepancy between evaluations from the reward model and GPT-4o-mini, indicating consistency across different evaluators.However, when compared to Direct Fine-tuning, our approach achieves a high win rate when evaluated by the reward model but falls below 50\% when evaluated by GPT-4o-mini.  This discrepancy arises because our classifier model is trained on data labeled by the reward model, meaning it inherently aligns with the reward model’s preferences. In contrast, Direct Fine-tuning leverages a fine-tuned model trained on reward model from GPT-4-preference data, leading to a stronger alignment with GPT-4o-mini's evaluation.

\paragraph{Two Preference Dimensions}
For two dimensional experiments, since there are in total 12 combinations of preferences, we randomly choose 8 out of these 12 combinations to reduce the cost when doing the evaluation. Similar to single preference dimension experiments, we use the same $(\alpha_1, \alpha_2)$ values across all experiments without tuning them for different preference combinations. While this approach may severely undermine the performance of our method, it serves as a stress test for its sensitivity to hyperparameters, making our evaluation more practical by assessing robustness rather than relying on delicate hyperparameter selection.
Table \ref{tab:2d_combined_results} shows the average win rate when combining 2 preference dimensions. Direct prompting remains a strong baseline, with our approach achieving only a 50.47\% win rate against it. However, compared to the vanilla baseline, our method consistently outperforms, attaining an average win rate of 72.45\%. While our approach does not show a clear advantage over direct prompting, this aligns with prior findings \cite{jang2023personalized}, which highlight direct prompting as a strong baseline. Nevertheless, direct prompting is limited to explicitly stated preferences and is not applicable when preferences are implicit. Compared to p-soup, our win rate remains below 50\% when evaluated by GPT-4o-mini. This is expected, as p-soup, while not directly fine-tuned on two-dimensional preference combinations, benefits from merging parameters from fine-tuned models. Given that only two dimensions are merged, it still achieves strong performance. Additionally, we observe that the win rate for Koala is generally higher than for Ultrafeedback, consistent with the results from single-dimensional experiments.
\begin{figure}[h]
    \centering
    \includegraphics[width=\linewidth]{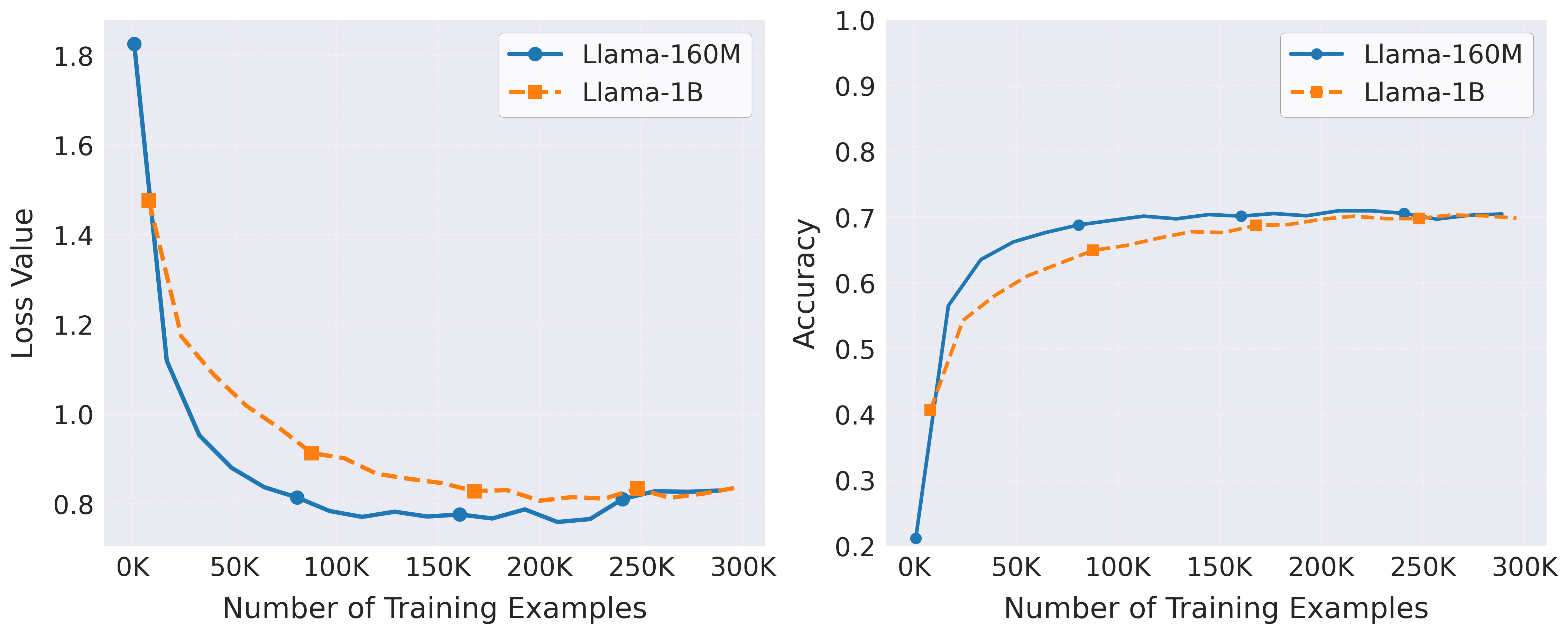}
    \caption{Accuracy and loss on evaluation set for classifier using Llama-160M and Llama-1B respectively. }
    \label{fig: eval loss and acc}
     \vspace{-0.5cm}
\end{figure}
\begin{figure}[h]
    \centering
    \includegraphics[width=\linewidth]{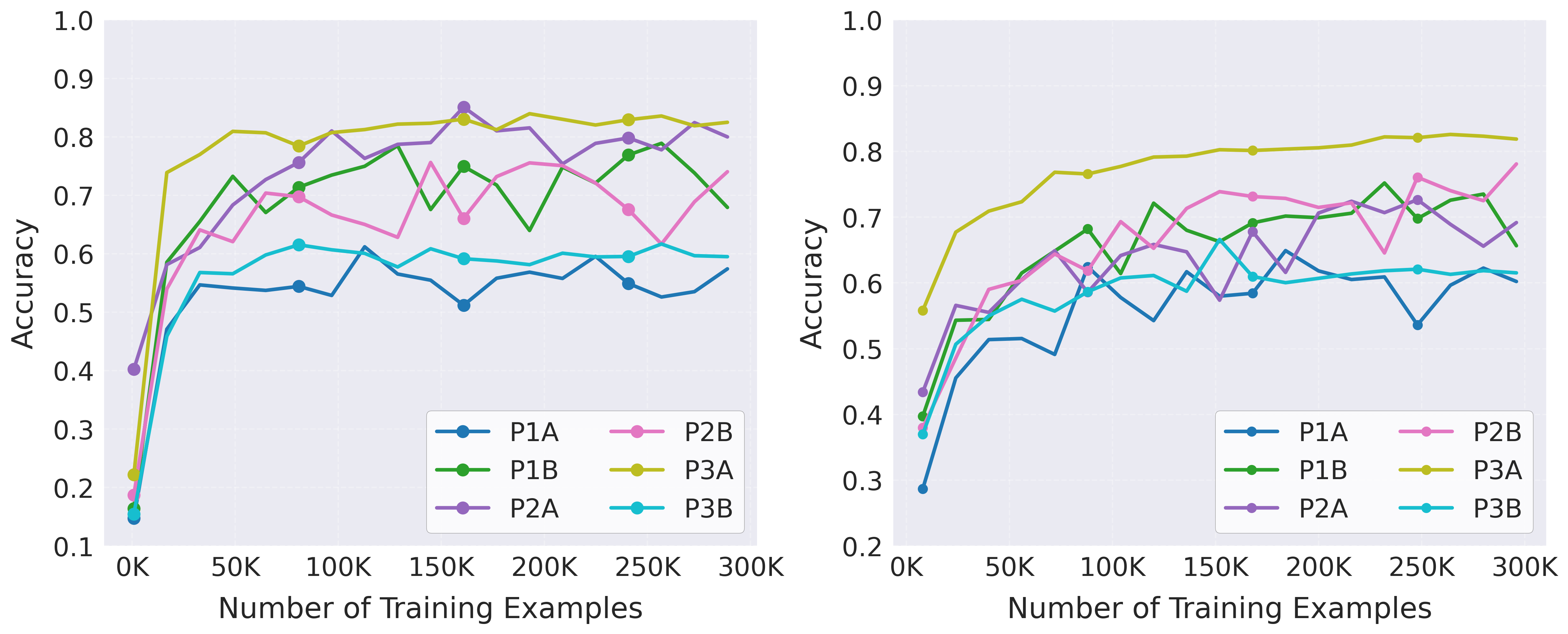}
    \vspace{-5ex}
    \caption{Per-dimension accuracy on evaluation dataset with Llama-160M and Llama-1B respectively. }
    \label{fig: per class accuracy}
     \vspace{-0.5cm}
\end{figure}

\begin{figure*}[h]
    \centering
\vspace{-0.5cm}\includegraphics[width=0.7\linewidth]{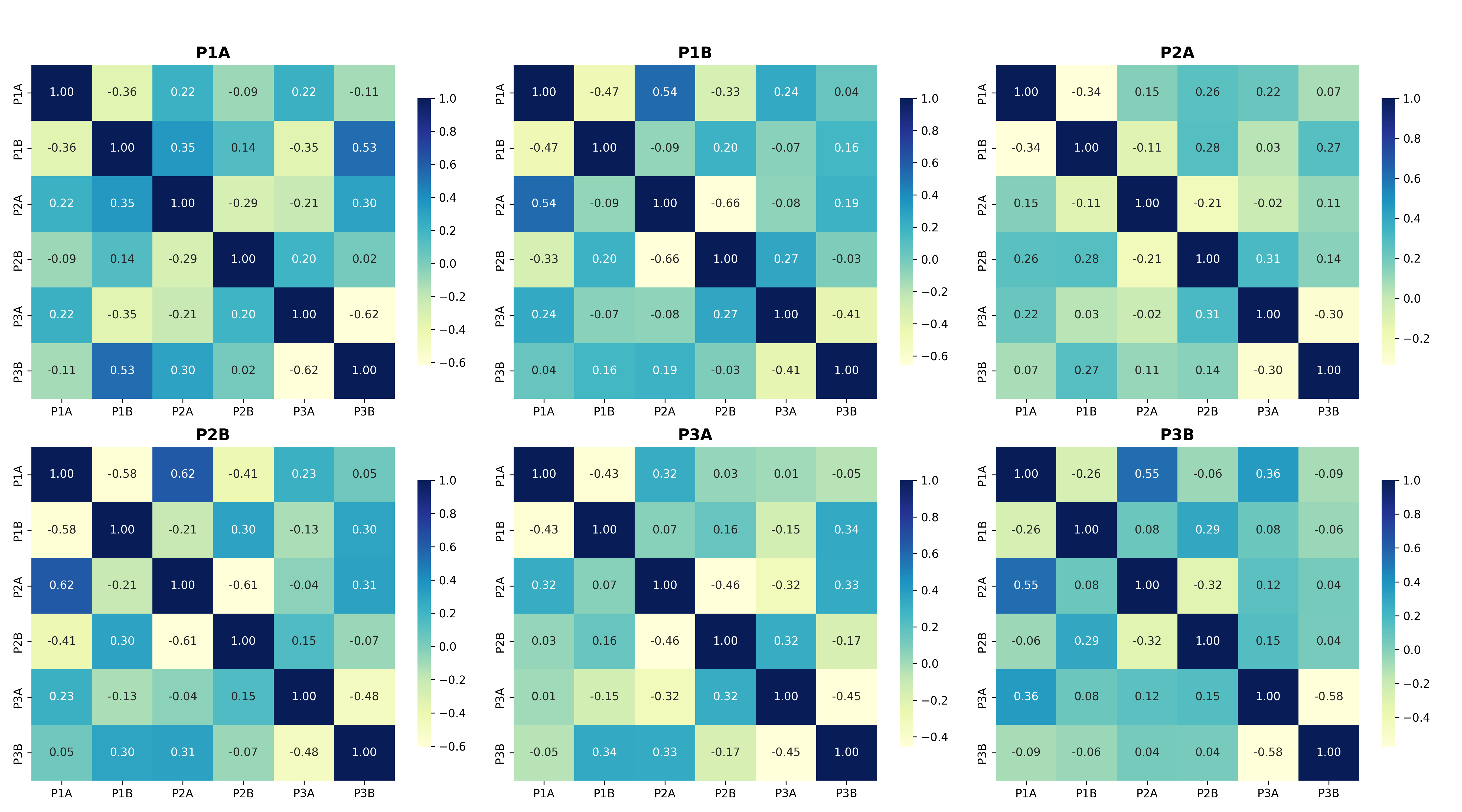}
    \caption{\textbf{Correlation matrices for reward values between different preference dimensions.} Each matrix only uses the text generated by direct prompting using the preference dimension shown in each title. For each matrix, the reward values for all preference dimensions are calculated and correlation calculation is performed. }
    \label{fig: correlation matrix}
     \vspace{-0.5cm}
\end{figure*}

\paragraph{Three Preference Dimensions} 
Table \ref{tab:3d_combined_results} presents the results for three-preference dimension combinations, where CLIPer consistently outperforms all three baselines in both reward model evaluation and GPT-4o-mini evaluation. Comparing these results with those from the single-dimensional setting (Table \ref{tab:combined_results}) and the two-dimensional setting (Table \ref{tab:2d_combined_results}), we observe a notable trend: the win rate of our approach, as evaluated by GPT-4o-mini, increases progressively from approximately 40\% in the single-dimensional case to above 45\% in the two-dimensional case and surpasses 50\% in the three-dimensional case. This trend suggests that while p-soup remains competitive at lower preference dimensions, its performance diminishes as the number of preference dimensions increases. In contrast, our approach maintains strong performance and even improves as the dimensionality scales, highlighting its robustness in handling more complex preference interactions. This observation underscores the limitations of parameter merging in p-soup when dealing with higher-dimensional preference combinations, whereas our method demonstrates better adaptability and effectiveness in capturing nuanced preference structures.
\begin{remark}
Note that the performance of our method could be better if we tune-our hyper-parameter more carefully, for example, using more in-domain evaluation data, or tune the hyper-parameter for different preference combinations. Interested readers can refer to Appendix \ref{app: more exp} to see more results using different hyper-parameters. 
\end{remark}
\subsection{Further Analysis}

\paragraph{Scaling up the Classifier Model}
While we use LLaMA-160M as our classifier in the main experiments, we conduct additional experiments to investigate whether increasing the model size further improves the performance of the classifier model. In Figure \ref{fig: eval loss and acc}, we compare the evaluation loss and accuracy when using LLaMA-160M and LLaMA-1B. Surprisingly, we observe no significant gap in classification performance in terms of loss and accuracy, suggesting that LLaMA-160M is sufficient for our current experiments. This indicates that scaling up the classifier model may not provide substantial benefits in this setting. In Figure \ref{fig: per class accuracy}, we further plot the Per-class accuracy for each class, we found that P3A,  which corresponds to "Playful", as in Table \ref{tab: preference prompt for each dimension}, has the lowest classification accuracy, while P1A, which corresponds to "elementary level", has the highest classification accuracy. Even the dimension has the lowest accuracy achieves more than 0.5 (over 0.17) classification accuracy, showing that the classifier is indeed classifying these classes. 

\paragraph{Correlation Analysis}
An important assumption for our approach is that the preference dimensions we are combining should be independent since during inference the classifier output probability is multiplied across preference dimensions. To empirically verify this, in Figure \ref{fig: correlation matrix}, we analyze the correlation of different dimensions where we generate text that is intentionally steered toward that dimension through direct prompting. For each text, we use the reward model to label each dimension, and compute the Pearson correlation coefficient over the reward model scores among all preference dimension pairs. We observe that for every preference and its opposite, the correlation is always very strongly negative. This is expected and not an issue since we never group opposite preferences in to $\textbf{c}$. Another observation is that when the text is generated based on a particular preference, it usually has low correlation with other preferences (except itself and the opposite preference), suggesting preference dimensions that we are combining are fairly independent from each other, which verifies our independence assumption.
\paragraph{Human Evaluation}
We get 6 PhD student volunteers to do human study, where we present them 50 questions and pairwise model response (either our response or the response from preference prompting, presented in randomized order) from the koala dataset, and ask them to evaluation which response satisfies “elementary level, playful and concise” dimensions better, the results are given in Table \ref{tab:human_pref_wins} and the agreement among the labelers are give in Table \ref{tab:agreement_majority}. In average, our approach holds a win rate of 65.3\% when evaluated by human.


\section{Conclusion}
In this paper, we introduced CLIPer, a classifier-guided inference-time personalization approach that enables dynamic adaptation of LLMs to diverse user preferences without  extensive fine-tuning. By leveraging a lightweight classifier model, CLIPer steers generation in real time, offering a scalable and efficient alternative to traditional fine-tuning approaches. Our experimental results demonstrate the potential of CLIPer as a practical and adaptable solution for real-world LLM personalization.

\section*{Limitations}
As our work focuses on building a scalable and efficient personalization framework—an intrinsically challenging task—it naturally introduces certain limitations. First, our classifier model is trained on synthetic data derived from reward model rankings, rather than human-annotated labels. While this design choice enables scalable training without costly human supervision, it may also introduce alignment biases from the reward model. However, our empirical evaluations across both in-domain (reward model) and out-of-domain (GPT-4o-mini) judges help mitigate this concern.
Second, our method assumes conditional independence between preference dimensions to avoid exponential scaling. While this assumption holds empirically for the dimensions we evaluate (as shown in our correlation analysis), future work could explore modeling interdependencies explicitly to support more nuanced personalization.
Third, due to resource constraints, we evaluate CLIPer on a fixed set of six well-defined preference dimensions. This allows for controlled comparisons and benchmarking, but does not fully capture the range of real-world user preferences—especially those that are implicit or context-specific. Nonetheless, the framework is general and can easily accommodate new dimensions as data becomes available.
Finally, we apply a shared set of hyperparameters across all combinations of preferences to ensure fairness and reduce tuning overhead. While this likely understates the upper bound of our method’s performance, it demonstrates robustness and practicality for real-world use without extensive hyperparameter tuning.

\bibliography{custom}

\appendix
\onecolumn
\section{Experimental details}\label{app: exp_detail}

\begin{figure*}[h]
    \centering
    \includegraphics[width=0.8\linewidth]{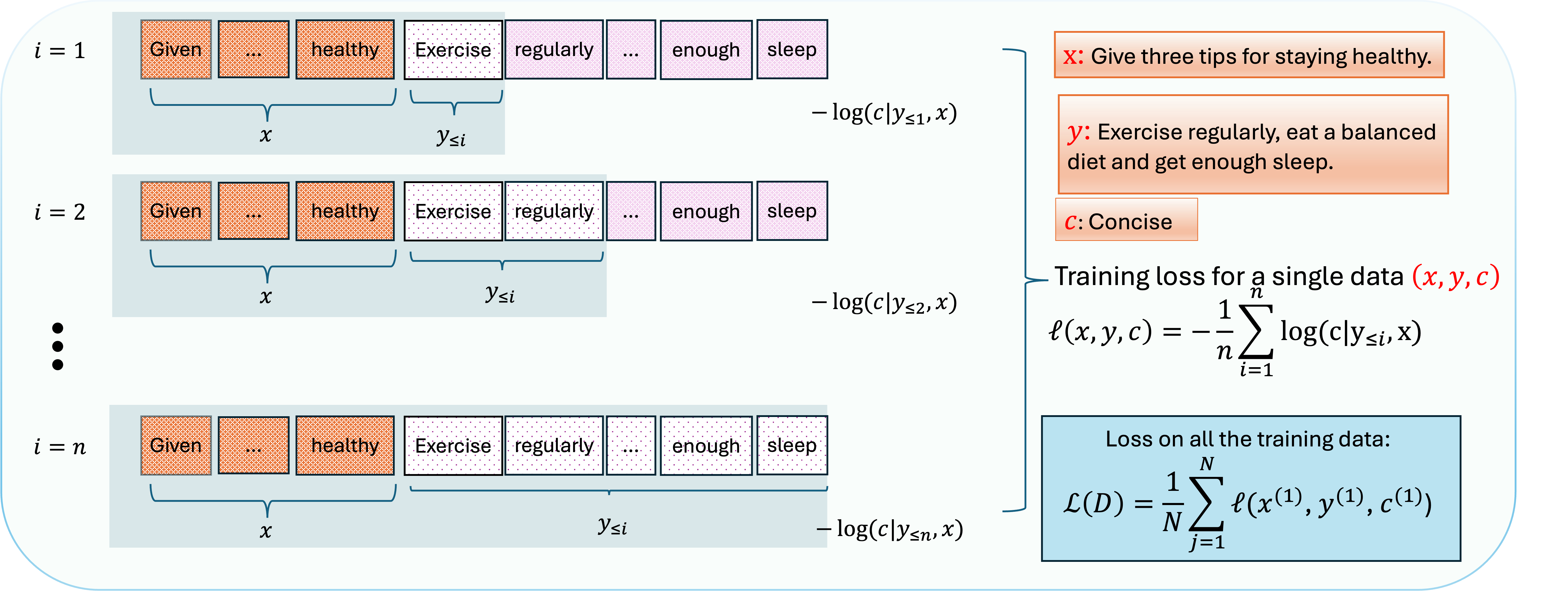}
    \caption{Illustration of the composition of the training loss using a single data point $(x, y,c)$, where $x$ is the user prompt, $y$ is the LLM generation, and $c$ is the preference that the texts fits into. Training loss is the average of the negative log likelihood of preference $c$ given the partial generation $y_{\leq i}$ and the prompt $x$.
And the overall loss function across the training dataset $D=\{(x^{(j)}, y^{(j)}, c^{(j)})\}$ is averaged over $N$ data. }
    \label{fig: loss} 
\end{figure*}

\paragraph{Parameter Setting for Classifier} For Llama-160M model, we search over learning rate being $\{1e-4, 1e-5, 1e-6\}$, and choose best learning rate based on the evaluation loss. The final learning rate used in our experiment is $1e-5$. For Llama-1B model, we also search over learning rate being  $\{1e-4, 1e-5, 1e-6\}$ and use the learning rate $1e-6$, which has the lowest evaluation loss.
\paragraph{Illustration of the loss}
The illustation of the composition of the loss is in Figure \ref{fig: loss}
\section{Additional experiments}\label{app: more exp}
\subsection{Single dimension}
\paragraph{Performance of Alpaca-Eval subsets}
We use two subsets randomly sampled from Alapaca-eval, each of 100 data prompts, and select the hyper-parameter $\alpha$ based on its win rate against direct prompting. Table \ref{tab: 1} (evaluated by GPT-4o-mini) and Table \ref{tab: 2} (evaluated by reward model) illustrate the win rate of different dimensions using different $\alpha$ on both subset 1 and subset 2. Table \ref{tab: 4} and Table \ref{tab: 5} summarized the average win rate across all 6 dimensions, evaluated on gpt-4o-mini and reward model, respectively. 
\begin{table}[ht]
\centering
\begin{tabular}{c|c|cccccc|c}
\toprule
\textbf{Dataset} & \textbf{$\alpha$} & \textbf{P1A} & \textbf{P1B} & \textbf{P2A} & \textbf{P2B} & \textbf{P3A} & \textbf{P3B} & \textbf{Avg.} \\
\midrule
\multirow{6}{*}{AlpacaEval Subset 1} 
 & $\alpha=$0.05 & 0.555 & 0.495 & 0.495 & 0.485 & 0.450 & 0.560 & 0.507\\\cline{2-9}
 &$\alpha=$ 0.1  & 0.490 & 0.525 & 0.500 & 0.485 & 0.495 & 0.500 & 0.499\\ \cline{2-9}
 & $\alpha=$0.3  & 0.540 & 0.535 & 0.555 & 0.565 & 0.465 & 0.535 & 0.533\\ \cline{2-9}
 & $\alpha=$0.5  & 0.530 & 0.515 & 0.495 & 0.480 & 0.535 & 0.590 & 0.524\\\cline{2-9}
 & $\alpha=$0.8  & 0.560 & 0.515 & 0.515 & 0.595 & 0.515 & 0.585 & \textbf{0.548}\\\cline{2-9}
 &$\alpha=$ 1.0  & 0.545 & 0.495 & 0.505 & 0.455 & 0.520 & 0.645 & 0.528\\
\midrule
\multirow{6}{*}{AlpacaEval Subset 2} 
 & $\alpha=$0.05 & 0.555 & 0.500 & 0.540 & 0.590 & 0.475 & 0.490 & 0.525\\\cline{2-9}
 &$\alpha=$ 0.1  & 0.505 & 0.490 & 0.415 & 0.525 & 0.470 & 0.530 & 0.489\\ \cline{2-9}
 &$\alpha=$ 0.3  & 0.495 & 0.555 & 0.495 & 0.525 & 0.550 & 0.515 & 0.523\\\cline{2-9}
 &$\alpha=$ 0.5  & 0.530 & 0.565 & 0.590 & 0.505 & 0.485 & 0.510 & \textbf{0.531}\\\cline{2-9}
 & $\alpha=$0.8  & 0.460 & 0.530 & 0.515 & 0.495 & 0.545 & 0.585 & 0.522\\\cline{2-9}
 &$\alpha=$ 1.0  & 0.490 & 0.470 & 0.505 & 0.525 & 0.465 & 0.525 & 0.497\\
\bottomrule
\end{tabular}
\caption{Win rate (evaluated with gpt4o-mini) of single dimension on AlpacaEval Subset 1 and AlpacaEval Subset 2 against Direct Prompting baseline with various $\alpha$. The \textbf{best} $\alpha$ (based on the average value) for each dataset is highlighted.}
\label{tab: 1}
\end{table}

\begin{table*}[ht]
\centering
\begin{tabular}{ccccccc}
\toprule
$\alpha$ & \textbf{0.05} & \textbf{0.1} & \textbf{0.3} & \textbf{0.5} & \textbf{0.8} & \textbf{1.0} \\
\midrule
\textbf{Alpaca Subset \#.1 } & 0.5067 & 0.4992 & 0.5325 & \textbf{\textcolor{mediumgreen}{0.5242}} & \textcolor{blue}{0.5475} & 0.5275 \\
\textbf{Alpaca Subset \#.2 } & 0.5250 & 0.4892 & 0.5225 & \textcolor{mediumgreen}{0.5308} & \textbf{\textcolor{blue}{0.5217}} & 0.4967 \\\hline
\textbf{Avg.} & 0.5159 & 0.4942 & 0.5275 & 0.5275 & \textbf{0.5346} & 0.5121 \\
\bottomrule
\end{tabular}
\caption{Average Win Rates (across 6 dimensions) for different $\alpha$ across Alpaca Subset \#.1 and Alpaca Subset \#.2 evaluated on GPT4o-mini. We highlight the win rate of $\alpha$ selected using  Alpaca Subset \#.1 as the evaluation set and Alpaca Subset \#.2 as the test set  with \textcolor{blue}{blue}, while the win rate selected using  Alpaca Subset \#.2 as the evaluation set and Alpaca Subset \#.1 as the test set  with \textcolor{mediumgreen}{green}. For both of them, we bold the test set. We also bold the result that has the largest average for these two sets, which corresponds to a parameter $\alpha$ that will be used for direct evaluation on koala and ultrafeedback dataset. }
\label{tab: 4}
\end{table*}

\begin{table*}[ht]
\centering
\begin{tabular}{ccccccc}
\toprule
$\alpha$ & \textbf{0.05} & \textbf{0.1} & \textbf{0.3} & \textbf{0.5} & \textbf{0.8} & \textbf{1.0} \\
\midrule
\textbf{Alpaca Subset \#.1 } & 0.5092 & 0.5183 & 0.5092 & 0.5275 &\cellcolor{highlightblue} {0.5408} &{0.4917} \\
\textbf{Alpaca Subset \#.2 }  & 0.5100 & 0.5033 & 0.5167 & 0.5317 & \cellcolor{highlightblue} {0.5408} & {0.5408}\\\hline
\textbf{Avg.} & 0.5096 & 0.5108 & 0.5130 & 0.5296 & \textbf{0.5408} & 0.5162 \\
\bottomrule
\end{tabular}
\caption{Average Win Rates (across 6 dimensions) for different $\alpha$ across Alpaca Subset 1 and Alpaca Subset 2 evaluated on reward models. The win rate of $\alpha$ selected using Alpaca Subset 1 as the evaluation set and Alpaca Subset 2 as the test set, and vice versa, are consistent, which is highlighted in blue. We select the parameter $\alpha$ by taking the average of these two sets. }
\label{tab: 5}
\end{table*}

\begin{table}[ht]
\centering
\begin{tabular}{c|c|cccccc|c}
\toprule
\textbf{Dataset} & \textbf{$\alpha$} & \textbf{P1A} & \textbf{P1B} & \textbf{P2A} & \textbf{P2B} & \textbf{P3A} & \textbf{P3B} & \textbf{Avg.} \\
\midrule
\multirow{6}{*}{AlpacaEval Subset 1} 
 & $\alpha=$ 0.05 & 0.590 & 0.500 & 0.465 & 0.475 & 0.450 & 0.575 & 0.509\\
 & $\alpha=$ 0.1  & 0.555 & 0.490 & 0.520 & 0.465 & 0.540 & 0.540 & 0.518\\
 & $\alpha=$ 0.3  & 0.530 & 0.525 & 0.535 & 0.430 & 0.530 & 0.505 & 0.509\\
 & $\alpha=$ 0.5  & 0.520 & 0.545 & 0.545 & 0.515 & 0.470 & 0.570 & 0.528\\
 & $\alpha=$ 0.8  & 0.605 & 0.515 & 0.505 & 0.520 & 0.550 & 0.550 & \textbf{0.541}\\
 & $\alpha=$ 1.0  & 0.510 & 0.490 & 0.500 & 0.435 & 0.490 & 0.525 & 0.492\\
\midrule
\multirow{6}{*}{AlpacaEval Subset 2} 
 & $\alpha=$ 0.05 & 0.490 & 0.495 & 0.460 & 0.590 & 0.495 & 0.530 & 0.510\\
 & $\alpha=$ 0.1  & 0.510 & 0.515 & 0.440 & 0.455 & 0.540 & 0.560 & 0.503\\
 & $\alpha=$ 0.3  & 0.505 & 0.585 & 0.455 & 0.470 & 0.545 & 0.540 & 0.517\\
 & $\alpha=$ 0.5  & 0.505 & 0.560 & 0.525 & 0.530 & 0.515 & 0.555 & 0.532\\
 & $\alpha=$ 0.8  & 0.575 & 0.545 & 0.495 & 0.530 & 0.515 & 0.585 & \textbf{0.541}\\
 & $\alpha=$ 1.0  & 0.500 & 0.560 & 0.535 & 0.455 & 0.555 & 0.640 & \textbf{0.541}\\
\bottomrule
\end{tabular}
\caption{Win rate (evaluated with reward model) of single dimension on AlpacaEval Subset 1 and AlpacaEval Subset 2 against Direct Prompting baseline with various $\alpha$. The \textbf{best} $\alpha$ for each dataset is highlighted.}
\label{tab: 2}
\end{table}

\paragraph{Performance on Koala and UltraFeedback dataset} Table \ref{tab: 3} illustrates the win rate against direct fine-tuning and direct prompting for each dimension and their average on Koala and UltraFeedback dataset, when evaluated by reward model. 
\begin{table}[ht]
\centering
\begin{tabular}{c|c|c|cccccc|c}
\toprule
Baseline&\textbf{Dataset} & \textbf{$\alpha$} & \textbf{P1A} & \textbf{P1B} & \textbf{P2A} & \textbf{P2B} & \textbf{P3A} & \textbf{P3B} & \textbf{Avg.} \\
\midrule
\multirow{12}{*}{p-soup \& Direct Fine-tuning}&\multirow{6}{*}{Koala} 
 & $\alpha=$ 0.05 & 0.82 & 0.74 & 0.92 & 0.94 & 0.92 & 0.24 & 0.765\\
& & $\alpha=$ 0.1  & 0.78 & 0.84 & 0.88 & 0.90 & 0.96 & 0.22 & 0.765\\
& & $\alpha=$ 0.3  & 0.80 & 0.86 & 0.88 & 0.88 & 0.88 & 0.24 & 0.757\\
& & $\alpha=$ 0.5  & 0.80 & 0.82 & 0.88 & 0.90 & 0.88 & 0.26 & 0.758\\
& & $\alpha=$ 0.8  & 0.82 & 0.88 & 0.90 & 0.96 & 0.90 & 0.26 & \textbf{0.873}\\
& & $\alpha=$ 1.0  & 0.80 & 0.86 & 0.88 & 0.88 & 0.88 & 0.28 & 0.798\\
\cline{2-10}
&\multirow{6}{*}{UltraFeedback} 
 & $\alpha=$ 0.05 & 0.76 & 0.68 & 0.76 & 0.74 & 0.78 & 0.24 & 0.660\\
& & $\alpha=$ 0.1  & 0.82 & 0.80 & 0.82 & 0.76 & 0.88 & 0.22 & 0.717\\
& & $\alpha=$ 0.3  & 0.76 & 0.76 & 0.78 & 0.80 & 0.80 & 0.26 & 0.693\\
& & $\alpha=$ 0.5  & 0.80 & 0.86 & 0.78 & 0.76 & 0.90 & 0.26 & \textbf{0.760}\\
& & $\alpha=$ 0.8  & 0.74 & 0.84 & 0.74 & 0.86 & 0.82 & 0.28 & 0.713\\
& & $\alpha=$ 1.0  & 0.80 & 0.84 & 0.78 & 0.80 & 0.82 & 0.26 & 0.717\\
\midrule
\multirow{12}{*}{Direct Prompting}& \multirow{6}{*}{Koala} 
 & $\alpha=$ 0.05 & 0.36 & 0.33 & 0.52 & 0.66 & 0.43 & 0.54 & 0.473\\
& & $\alpha=$ 0.1  & 0.51 & 0.38 & 0.48 & 0.51 & 0.38 & 0.55 & 0.468\\
& & $\alpha=$ 0.3  & 0.43 & 0.46 & 0.58 & 0.55 & 0.34 & 0.49 & 0.475\\
& & $\alpha=$ 0.5  & 0.52 & 0.51 & 0.45 & 0.60 & 0.32 & 0.51 & 0.468\\
& & $\alpha=$ 0.8  & 0.58 & 0.52 & 0.43 & 0.65 & 0.44 & 0.60 & 0.537\\
& & $\alpha=$ 1.0  & 0.54 & 0.58 & 0.53 & 0.51 & 0.44 & 0.63 & \textbf{0.538}\\
\cline{2-10}
&\multirow{6}{*}{UltraFeedback} 
 & $\alpha=$ 0.05 & 0.46 & 0.34 & 0.45 & 0.43 & 0.44 & 0.46 & 0.430\\
& & $\alpha=$ 0.1  & 0.40 & 0.39 & 0.54 & 0.51 & 0.38 & 0.45 & 0.445\\
& & $\alpha=$ 0.3  & 0.54 & 0.44 & 0.52 & 0.53 & 0.38 & 0.54 & 0.493\\
& & $\alpha=$ 0.5  & 0.54 & 0.56 & 0.53 & 0.62 & 0.43 & 0.55 & \textbf{0.538}\\
& & $\alpha=$ 0.8  & 0.60 & 0.54 & 0.50 & 0.51 & 0.46 & 0.61 & 0.537\\
& & $\alpha=$ 1.0  & 0.45 & 0.57 & 0.46 & 0.45 & 0.44 & 0.51 & 0.480\\
\bottomrule
\end{tabular}
\caption{Win rate against P-soup and Direct Prompting (evaluated with reward model) of single dimension on the Koala and UltraFeedback datasets with various $\alpha$. The \textbf{best} $\alpha$ for each dataset is highlighted.}
\label{tab: 3}
\end{table}
\subsection{Two Dimensions}
See Table \ref{tab: 6} and Table \ref{tab: 7} for alpaca subset results  while the Koala and ultrafeedback results in Table \ref{tab: 8} and \ref{tab: 9}. 
\subsection{Three dimensions}
See Table \ref{tab: 10} and Table \ref{tab: 11} for alpaca subset results  while the Koala and ultrafeedback results in Table \ref{tab: 12} and \ref{tab: 13}. 
\begin{table}[ht]
\centering
\begin{tabular}{llrrrr}
\toprule
&$(\alpha_1, \alpha_2)$ &  (0.50, 0.50) &  (0.80, 0.50) &  (0.50, 0.80) &  (0.80, 0.80) \\
\midrule
\multirow{2}{*}{Direct Prompting}&Koala  & 0.5394 & 0.5275 & 0.5169 & 0.5175 \\
&UltraFeedback  & 0.4913 & 0.4844 & 0.5119 & 0.4825 \\
\multirow{2}{*}{p-soup \& Direct Fine-tuning}&Koala  & 0.7200 & 0.7063 & 0.7125 & 0.7025 \\
&UltraFeedback  & 0.6556 & 0.6550 & 0.6587 & 0.6438 \\
\multirow{2}{*}{Vanilla Baseline}&Koala  & 0.7725 & 0.7775 & 0.7612 & 0.7662 \\
&UltraFeedback  & 0.7100 & 0.7125 & 0.7200 & 0.7056 \\
\bottomrule
\end{tabular}
\caption{Reward model evaluation result on the Koala and UltraFeedback datasets, where we choose $\alpha_1, \alpha_2\in \{0.5, 0.8\}$.} 
\label{tab: 8}
\end{table}
\begin{table}[ht]
\centering
\begin{tabular}{llrrrr}
\toprule
&$(\alpha_1, \alpha_2)$ &  (0.50, 0.50) &  (0.80, 0.50) &  (0.50, 0.80) &  (0.80, 0.80) \\
\midrule
\multirow{2}{*}{Direct Prompting} &Koala  & 0.5031 & 0.5212 & 0.5350 & 0.5262 \\
&UltraFeedback  & 0.5038 & 0.4938 & 0.5112 & 0.4900 \\
\multirow{2}{*}{p-soup \& Direct Fine-tuning}&Koala  & 0.4775 & 0.4644 & 0.4963 & 0.4694 \\
&UltraFeedback  & 0.4537 & 0.4437 & 0.4469 & 0.4394 \\ 
\multirow{2}{*}{Vanilla Baseline}&Koala  & 0.7106 & 0.6987 & 0.7244 & 0.6987 \\
&UltraFeedback  & 0.6975 & 0.6981 & 0.6956 & 0.6925 \\
\bottomrule
\end{tabular}
\caption{GPT-4o-mini evaluation result on the Koala and UltraFeedback datasets, where we choose $\alpha_1, \alpha_2\in \{0.5, 0.8\}$.} 
\label{tab: 9}
\end{table}
\begin{table}[ht]
\scriptsize
\centering
\begin{tabular}{p{2.5cm}p{2cm}p{1cm}p{1cm}p{1cm}p{1cm}p{1cm}p{1cm}p{1cm}p{1cm}}
\toprule
& $(\alpha_1, \alpha_2, \alpha_3)$ & (0.5, 0.5, 0.5) & (0.8, 0.8, 0.5) & (0.8, 0.5, 0.5) & (0.5, 0.8, 0.8) & (0.5, 0.8, 0.5) & (0.5, 0.5, 0.8) & (0.8, 0.5, 0.8) & (0.8, 0.8, 0.8) \\
\midrule
\multirow{2}{*}{Direct Prompting} & Koala  & 0.6000 & 0.5533 & 0.5933 & 0.5333 & 0.5600 & 0.5700 & 0.5933 & 0.5733 \\
& UltraFeedback  & 0.5467 & 0.5333 & 0.5300 & 0.5367 & 0.4700 & 0.5600 & 0.4600 & 0.4767 \\
\multirow{2}{*}{p-soup \& Direct Fine-tuning} & Koala  & 0.8400 & 0.8667 & 0.8533 & 0.8333 & 0.8133 & 0.8467 & 0.8533 & 0.8267 \\
& UltraFeedback  & 0.8133 & 0.8267 & 0.8067 & 0.8200 & 0.7533 & 0.8133 & 0.7533 & 0.7467 \\
\multirow{2}{*}{Vanilla Baseline} & Koala  & 0.8800 & 0.9067 & 0.8667 & 0.8467 & 0.8800 & 0.8733 & 0.9000 & 0.8733 \\
& UltraFeedback  & 0.8533 & 0.8267 & 0.8267 & 0.8400 & 0.8033 & 0.8400 & 0.7900 & 0.7767 \\
\bottomrule
\end{tabular}
\caption{Reward model evaluation result on the Koala and UltraFeedback datasets in the 3D setting, where we choose $\alpha_1, \alpha_2, \alpha_3 \in \{0.5, 0.8\}$.} 
\label{tab: 12}
\end{table}
\begin{table}[ht]
\scriptsize
\centering
\begin{tabular}{p{2.5cm}p{2cm}p{1cm}p{1cm}p{1cm}p{1cm}p{1cm}p{1cm}p{1cm}p{1cm}}
\toprule
& $(\alpha_1, \alpha_2, \alpha_3)$ & (0.8, 0.5, 0.8) & (0.8, 0.5, 0.5) & (0.5, 0.5, 0.8) & (0.5, 0.5, 0.5) & (0.5, 0.8, 0.8) & (0.5, 0.8, 0.5) & (0.8, 0.8, 0.5) & (0.8, 0.8, 0.8) \\
\midrule
\multirow{2}{*}{Direct Prompting} & Koala  & 0.5133 & 0.5000 & 0.5067 & 0.5367 & 0.5100 & 0.4867 & 0.5267 & 0.5000 \\
& UltraFeedback  & 0.4600 & 0.5000 & 0.5100 & 0.5367 & 0.4767 & 0.4633 & 0.4967 & 0.5000 \\
\multirow{2}{*}{p-soup \& Direct Fine-tuning} & Koala  & 0.4900 & 0.5600 & 0.5700 & 0.5467 & 0.5533 & 0.5667 & 0.5333 & 0.5033 \\
& UltraFeedback  & 0.5467 & 0.5600 & 0.5833 & 0.5833 & 0.5533 & 0.5200 & 0.5633 & 0.5700 \\
\multirow{2}{*}{Vanilla Baseline} & Koala  & 0.7867 & 0.7900 & 0.7833 & 0.8067 & 0.7967 & 0.7900 & 0.8100 & 0.7833 \\
& UltraFeedback  & 0.7500 & 0.7700 & 0.8033 & 0.8100 & 0.7800 & 0.7633 & 0.7967 & 0.7633 \\
\bottomrule
\end{tabular}
\caption{GPT-4o-mini evaluation result on the Koala and UltraFeedback datasets in the 3D setting, where we choose $\alpha_1, \alpha_2, \alpha_3 \in \{0.5, 0.8\}$.} 
\label{tab: 13}
\end{table}

\begin{table}[ht]
\centering
\begin{tabular}{lrrrr}
\toprule
$(\alpha_1, \alpha_2)$ &  (0.50, 0.50) &  (0.80, 0.50) &  (0.50, 0.80) &  (0.80, 0.80) \\
\midrule
Subset 1 &        0.5222 &        0.5141 &        0.5078 &        0.5000 \\
Subset 2 &        0.5103 &        0.5084 &        0.5006 &        0.5000 \\\hline
Average  &        0.5162 &        0.5112 &        0.5042 &        0.5000 \\
\bottomrule
\end{tabular}
\caption{gpt4o-mini evaluation result for Alpaca-eval subsets, where we choose $\alpha_1, \alpha_2\in \{0.5, 0.8\}$}.
\label{tab: 6}
\end{table}
\begin{table}[ht]
\centering
\begin{tabular}{lrrrr}
\toprule
$(\alpha_1, \alpha_2)$ &  (0.50, 0.50) &  (0.80, 0.50) &  (0.50, 0.80) &  (0.80, 0.80) \\
\midrule
Subset 1  & 0.5228 & 0.5409 & 0.5228 & 0.5112 \\
Subset 2  & 0.5012 & 0.5188 & 0.5166 & 0.5047 \\\hline
Average  & 0.5120 & 0.5298 & 0.5197 & 0.5080 \\
\bottomrule
\end{tabular}
\caption{Reward model evaluation result for Alpaca-eval subsets with reward model, where we choose $\alpha_1, \alpha_2\in \{0.5, 0.8\}$.} 
\label{tab: 7}
\end{table}

\begin{table*}[ht]
\centering
\scriptsize
\begin{tabular}{lrrrrrrrr}
\toprule
$(\alpha_1, \alpha_2, \alpha_3)$ & (0.5, 0.5, 0.5) & (0.8, 0.8, 0.5) & (0.8, 0.5, 0.5) & (0.5, 0.8, 0.8) & (0.5, 0.8, 0.5) & (0.5, 0.5, 0.8) & (0.8, 0.5, 0.8) & (0.8, 0.8, 0.8) \\
\midrule
Subset 1  & 0.5183 & 0.4800 & 0.5000 & 0.5017 & 0.5000 & 0.4833 & 0.5083 & 0.5000 \\
Subset 2  & 0.5250 & 0.5100 & 0.5000 & 0.5067 & 0.4983 & 0.4750 & 0.5317 & 0.5000 \\\hline
Average  & 0.5217 & 0.4950 & 0.5000 & 0.5042 & 0.4992 & 0.4792 & 0.5200 & 0.5000 \\
\bottomrule
\end{tabular}
\caption{GPT-4o-mini evaluation result for Alpaca-eval subsets in the 3D setting, where we choose $\alpha_1, \alpha_2, \alpha_3 \in \{0.5, 0.8\}$.} 
\label{tab: 10}
\end{table*}
\begin{table*}[ht]
\centering
\scriptsize
\begin{tabular}{lrrrrrrrr}
\toprule
$(\alpha_1, \alpha_2, \alpha_3)$ & (0.5, 0.5, 0.5) & (0.8, 0.8, 0.5) & (0.8, 0.5, 0.5) & (0.5, 0.8, 0.8) & (0.5, 0.8, 0.5) & (0.5, 0.5, 0.8) & (0.8, 0.5, 0.8) & (0.8, 0.8, 0.8) \\
\midrule
Subset 1  & 0.5017 & 0.5483 & 0.5217 & 0.5083 & 0.5167 & 0.5417 & 0.5617 & 0.5550 \\
Subset 2  & 0.5083 & 0.5383 & 0.4833 & 0.5117 & 0.5067 & 0.4933 & 0.5350 & 0.4850 \\\hline
Average  & 0.5050 & 0.5433 & 0.5025 & 0.5100 & 0.5117 & 0.5175 & 0.5483 & 0.5200 \\
\bottomrule
\end{tabular}
\caption{Reward model evaluation result for Alpaca-eval subsets in the 3D setting, where we choose $\alpha_1, \alpha_2, \alpha_3 \in \{0.5, 0.8\}$.} 
\label{tab: 11}
\end{table*}
\section{Human Evaluation}
In Table \ref{tab:human_pref_wins}, we show small scale human evaluation with 6 PhD students. Table \ref{tab:human_pref_wins} shows the inter-rater consistency measured as agreement rate with the majority vote across the 6 labelers. 
\begin{table}[t]
\centering
\small
\begin{tabular}{lccc}
\toprule
\textbf{Labeler} & \textbf{Our Wins} & \textbf{Preference Prompting Wins} & \textbf{Our Win Rate} \\
\midrule
Labeler 1 & 31 & 19 & 62\% \\
Labeler 2 & 43 & 17 & 86\% \\
Labeler 3 & 24 & 26 & 48\% \\
Labeler 4 & 25 & 25 & 50\% \\
Labeler 5 & 45 & 5  & 90\% \\
Labeler 6 & 28 & 22 & 56\% \\
\midrule
\textbf{Total} & \textbf{196} & \textbf{114} & \textbf{65.3\%} \\
\bottomrule
\end{tabular}
\caption{Human preference study with 6 PhD student volunteers. Each labeler evaluated 50 Koala questions with pairwise responses (ours vs.\ preference prompting) shown in randomized order, selecting which better satisfies the ``elementary-level, playful, and concise'' criteria.}
\label{tab:human_pref_wins}
\end{table}

\begin{table}[t]
\centering
\small
\begin{tabular}{lc}
\toprule
\textbf{Labeler} & \textbf{Agreement with Majority} \\
\midrule
Labeler 1 & 78\% \\
Labeler 2 & 92\% \\
Labeler 3 & 58\% \\
Labeler 4 & 50\% \\
Labeler 5 & 90\% \\
Labeler 6 & 70\% \\
\midrule
\textbf{Average} & \textbf{73.0\%} \\
\bottomrule
\end{tabular}
\caption{Inter-rater consistency measured as agreement rate with the majority vote across the 6 labelers.}
\label{tab:agreement_majority}
\end{table}
\clearpage
\newpage
\section{Prompt used for LLM as a Judge}
In Table \ref{tab:p1a}, we show prompt used to evaluate the wining response for preference P1A (elementary level) using LLM as a judge. Prompts for other preference dimensions can be found in the code for paper \cite{jang2023personalized} as we use the same prompt,
\begin{longtable}{|p{0.9\textwidth}|}
\hline
\hline
\textbf{Prompt for Evaluating P1A (Elementary Level)} \\
\hline
\endfirsthead
\hline
\endhead

\hline
\endfoot

\hline
\endlastfoot

You are a helpful following assistant whose goal is to select the response that meet the following preference criteria: \\

Choose a response that can easily be understandable by an elementary school student. \\

You should evaluate the responses only based on the provided preference criteria instead of the correctness or the helpfulness of the responses. \\

If both responses demonstrate the preference criteria reasonably well, it is a TIE. \\

If neither of the response demonstrates the preference criteria, it is a TIE. \\

\textbf{Annotation Guideline} \\

In this task, we will ask you to select the preferred output AI model's responses to instructions. \\

You will read a batch of examples, which are composed of the following: \\

1. an Instruction we give to the AI system \\
2. Output (a) \\
3. Output (b) \\

\textbf{Annotation Example} \\

For the annotation task, you will be asked to annotate an example. I.e. you will read all the examples first and then need to select the preferred output. Here's an example of how you should be formatting your answers for a batch of 3 examples. \\

\textbf{Example 1} \\

\textbf{Instruction 1:} \\

explain TypeScript and Duck Typing \\

\textbf{Output (a) for example 1:} \\

Hello, eager learner! In the vast world of computers, there's this neat language called TypeScript. Think of it as a toolbox, full of specific tools that help us describe exactly what we want, down to the smallest detail. And then there's this playful concept named Duck Typing. Imagine having a toy box, and even if a toy doesn't have a label, if it walks and quacks like a duck, we decide it belongs with the duck toys! It's all about how it acts. \\

\textbf{Output (b) for example 1:} \\

TypeScript: a statically-typed JavaScript variant. Duck Typing: dynamic type-checking. Essential knowledge. \\

\textbf{Thought for example 1:} \\

Output (a) uses relatable analogies like toolboxes, toy boxes, and ducks to explain the concepts, making it more accessible for younger minds. So the answer is Output (a). \\

\textbf{Result for example 1:} \\

Output (a) \\

\textbf{Example 2} \\

\textbf{Instruction 2:} \\

assume the persona of a therapist and walk me through how to be less reactive and use a pause ritual \\

\textbf{Output (a) for example 2:} \\

To systematically mitigate inherent reactivity: \\

Diaphragmatic Respiration: Activate the parasympathetic system and attenuate amygdala responses. \\

Somatic Anchoring: Establish a present-moment orientation, nullifying impulsive limbic reactions. \\

Cognitive Deliberation: Engage the dorsolateral prefrontal cortex for a nuanced appraisal of stimuli, curtailing reflexive emotional outputs. \\

Ensure consistent application for tangible results. \\

\textbf{Output (b) for example 2:} \\

Engaging in a cerebral exploration, are we? Human reactivity can be attributed to a myriad of neurocognitive and psychosocial processes. To temper this: \\

Diaphragmatic Breathing: This not only regulates physiological arousal but also modulates the amygdala's hypersensitivity. \\

Somatic Grounding: By tethering oneself to the present through sensory awareness, one can bypass the automatic, limbic-driven responses. \\

Cognitive Reflection: Ponder on the higher-order executive functions, allowing the prefrontal cortex to mediate the emotional reactions. \\

Regularly revisiting these strategies can aid in crafting a more poised persona. \\

\textbf{Thought for example 2:} \\

Both Output (a) and Output (b) use professional terms and complex, meaning that both of them are not easily understandable by an elementary school student. So the answer is TIE. \\

\textbf{Result for example 2:} \\

TIE \\

\textbf{Example 3} \\

\textbf{Instruction 3:} \\

Define a new language that contains words that are with numbers and letters such as 'IL9op'. \\

\textbf{Output (a) for example 3:} \\

A new language uses letters and numbers together. Like 'B2e' or 'M8te'. Understand? \\

\textbf{Output (b) for example 3:} \\

We're talking about a language where you combine letters and numbers to form words. Example: Instead of "cake", it's "C4k3". It's a straightforward concept. Get it? \\

\textbf{Thought for example 3:} \\

Both outputs are easily understandable by an elementary school student, but Output (b) might be a tad more understandable because of the direct comparison it offers ("Instead of 'cake', it's 'C4k3'"). So, the answer is Output (b). \\

\textbf{Result for example 3:} \\

Output (b) \\

\textbf{Annotation starts below} \\

Now is your turn. I will give you a fourth example. Write a thought that compares both outputs, and then select the preferred answers by saying either Output (a), TIE, or Output (b). \\

\textbf{Example 4} \\

\textbf{Instruction 4:} \\

\{instruction\} \\

\textbf{Output (a) for example 4:} \\

\{output\_1\} \\

\textbf{Output (b) for example 4:} \\

\{output\_2\} \\

\textbf{Preferred Output for example 4:} \\

Now write your thought and choose the preferred output for example 4. Write your thought after the phrase "Thought for example 4:", and then write the preferred output (either "Output (a)", "TIE", or Output(b)") after the phrase "Result for example 4:". \\

The preference criteria should favor responses that can easily be understandable by an elementary school student. You should evaluate the responses only based on the provided preference criteria instead of the correctness or the helpfulness of the responses. If both responses demonstrate the preference criteria reasonably well, it is a TIE. If neither of the response demonstrates the preference criteria, it is a TIE. \\

\textbf{Thought for Example 4:} \\
\end{longtable}
\captionof{table}{Prompt used to evaluate the wining response for preference P1A (elementary level) using LLM as a judge.} \label{tab:p1a}

\clearpage
\newpage
\section{Example of the Generation}

\begin{longtable}{|p{0.12\textwidth}|p{0.83\textwidth}|}
\hline
\textbf{\textit{Instruction}} & \textit{List the layers of the TCP/IP model and for each layer identify one protocol.} \\
\hline
\endfirsthead

\hline
\textbf{Preference Dimension} & \textbf{Output} \\
\hline
\endhead

\hline
\endfoot

\hline
\endlastfoot
\textbf{Preference Dimension} & \textbf{Output} \\

\hline

P1A & The TCP/IP model is like a cake! The cake is like the OSI model, which is a way to explain how computers communicate with each other. The OSI model has seven layers: 1. Physical Layer 2. Data Link Layer 3. Network Layer 4. Transport Layer 5. Session Layer 6. Presentation Layer 7. Application Layer. The Physical Layer is like the cake's crust. It's the part that holds everything together and makes sure the cake is safe to eat. The Data Link Layer is like the frosting on the cake. It's the part that makes the cake look pretty and delicious. The Network Layer is like the cake's middle layer. It's the part that helps the cake stay together and move around. The Transport Layer is like the cake's topping. It's the part that makes the cake taste good and enjoyable to eat. The Session Layer is like the candles on the cake. It's the part that helps the cake celebrate special occasions. The Presentation Layer is like the cake's decorations. It's the part that makes the cake look even more special and fun to eat. The Application Layer is like the cake's party. It's the part that helps the cake have a good time and make everyone happy. Just like how the cake has different layers that work together to make it delicious, the TCP/IP model has different layers that work together to make computers communicate with each other. \\
\hline

P1B & The TCP/IP model is a conceptual framework that describes the structure and function of the Internet. It consists of four layers: the application layer, the transport layer, the internet layer, and the link layer. At the application layer, one protocol is the Hypertext Transfer Protocol (HTTP). HTTP is a protocol that is used for transmitting information between a client and a server. It is used for tasks such as retrieving web pages, sending and receiving email, and streaming media. At the transport layer, one protocol is the Transmission Control Protocol (TCP). TCP is a protocol that is used for reliable data transmission. It ensures that data is sent in the proper order and that it is received without errors. At the internet layer, one protocol is the Internet Protocol (IP). IP is a protocol that is used for routing data packets between networks. It is responsible for determining the best path for data to take as it travels from its source to its destination. At the link layer, one protocol is the Ethernet. Ethernet is a protocol that is used for transmitting data over a physical network. It is the most commonly used protocol for wired networks and is used in both home and business environments. In summary, the TCP/IP model consists of four layers: the application layer, the transport layer, the internet layer, and the link layer. Each layer plays a critical role in the functioning of the Internet and is responsible for different aspects of data transmission and communication. \\
\hline

P2A & TCP/IP model consists of 4 layers: 1. Application layer 2. Transport layer 3. Internet layer 4. Network access layer. Protocols: 1. Application layer: HTTP, FTP, SMTP, DNS, DHCP, etc. 2. Transport layer: TCP, UDP, SCTP, etc. 3. Internet layer: IP, ICMP, ARP, etc. 4. Network access layer: Ethernet, Wi-Fi, PPP, etc. \\
\hline

P2B & The TCP/IP model is a conceptual framework used to understand how data is transmitted over a network. It consists of four layers: the application layer, the transport layer, the internet layer, and the link layer. 1. Application Layer: This is the top layer of the TCP/IP model and is responsible for creating, sending, and receiving data. Some common protocols used in this layer include HTTP, FTP, and SMTP. 2. Transport Layer: This layer is responsible for providing reliable data transfer between applications. Some common protocols used in this layer include TCP and UDP. 3. Internet Layer: This layer is responsible for routing data packets between networks. Some common protocols used in this layer include IPv4 and IPv6. 4. Link Layer: This is the lowest layer of the TCP/IP model and is responsible for providing physical connectivity between devices. Some common protocols used in this layer include Ethernet and Wi-Fi. In summary, the TCP/IP model is a framework that helps us understand how data is transmitted over a network. It consists of four layers, each with its own specific function, and different protocols are used at each layer to ensure efficient and reliable data transmission. \\
\hline

P3A & Okay, let's get down to business. The layers of the TCP/IP model are like a party, and each layer is a different guest. Here's the rundown: 1. Physical Layer: This is the hostess of the party. She's in charge of making sure everyone's connected to the right wires and cables. She's like Ethernet, always making sure everything's plugged in and working properly. 2. Data Link Layer: This is the guy who makes sure everyone's talking to each other. He's like Wi-Fi, always making sure the signals are clear and strong. 3. Network Layer: This is the party animal. He's like the Internet, always making sure everyone's connected and having a good time. 4. Transport Layer: This is the chill dude who makes sure everyone's getting along. He's like TCP, always making sure the data is sent and received correctly. 5. Session Layer: This is the party planner. She's like SIP, always making sure everyone's connected and coordinating the party. 6. Presentation Layer: This is the fashionista of the party. She's like HTTP, always making sure everyone's looking good and presenting themselves in the best way possible. 7. Application Layer: This is the party animal. He's like the app, always making sure everyone's having a good time and connecting with each other. So there you have it, folks! The layers of the TCP/IP model are like a party, and each layer is a different guest. And remember, when it comes to networking, it's all about having a good time and connecting with each other! \\
\hline

P3B & The layers of the TCP/IP model are: 1. Physical Layer: This layer is responsible for the physical transmission of data over a network. One protocol in this layer is Ethernet. 2. Data Link Layer: This layer is responsible for the reliable transmission of data between two devices on a network. One protocol in this layer is the Internet Protocol (IP). 3. Network Layer: This layer is responsible for routing data packets from their source to their destination. One protocol in this layer is the Internet Protocol (IP). 4. Transport Layer: This layer is responsible for end-to-end communication between applications. One protocol in this layer is the Transmission Control Protocol (TCP). 5. Session Layer: This layer is responsible for establishing, maintaining, and terminating connections between applications. One protocol in this layer is the Session Control Protocol (SCP). 6. Presentation Layer: This layer is responsible for data representation and encryption. One protocol in this layer is the Secure Sockets Layer (SSL). 7. Application Layer: This layer is responsible for user-level applications and protocols. One protocol in this layer is the Hypertext Transfer Protocol (HTTP). \\

\end{longtable}

\captionof{table}{Example 1: \textit{"List the layers of the TCP/IP model and for each layer identify one protocol."} with responses on different preference dimensions.} \label{tab:example}

\begin{longtable}{|p{0.12\textwidth}|p{0.83\textwidth}|}
\hline
\textbf{\textit{Instruction}} & \textit{Generate a recommendation for someone looking to buy a laptop.} \\
\hline
\endfirsthead

\hline
\textbf{Preference Dimension} & \textbf{Output} \\
\hline
\endhead

\hline
\endfoot

\hline
\endlastfoot

\textbf{Preference Dimension} & \textbf{Output} \\
\hline

P1A & If you're looking to buy a new laptop, I would recommend getting one that is lightweight, portable, and has a long battery life. A good laptop should have at least 8 hours of battery life so you can use it for a whole day without needing to charge it. It should also have a fast processor and plenty of memory so you can run multiple programs at once without slowing down. And of course, you want a screen that is big enough to see everything clearly, but not so big that it's heavy and hard to carry around. \\
\hline

P1B & As a PhD student in the field of computer science, I would recommend that you consider several factors when choosing a laptop. Firstly, you should consider the processing power and memory of the laptop. A laptop with a powerful processor and sufficient memory will be able to handle intensive tasks such as running complex simulations or processing large datasets. Additionally, you should consider the storage capacity of the laptop, as this will affect how much data you can store on the device. Secondly, you should consider the screen size and resolution of the laptop. A larger screen with a high resolution will provide a better viewing experience, particularly if you plan to use the laptop for tasks such as video editing or gaming. Finally, you should also consider the battery life of the laptop. A laptop with a long battery life will be more convenient for tasks that require you to be mobile, such as attending meetings or working from a coffee shop. Overall, when choosing a laptop, it is important to consider the specific tasks you will be performing and the environment in which you will be using the device. A laptop with a powerful processor, sufficient memory, a large screen with high resolution, and a long battery life will be well-suited for most computer science tasks. \\
\hline

P2A & If you're in the market for a new laptop, I recommend considering a lightweight and portable model with a long-lasting battery life. Look for a laptop with a fast processor, ample storage, and a high-resolution display for clear and vibrant visuals. Make sure to choose a laptop with a reliable brand and good warranty coverage for peace of mind. \\
\hline

P2B & When it comes to buying a laptop, there are several factors to consider. First, you need to determine your budget. Laptops can range from affordable options to high-end models that cost thousands of dollars. Next, you need to consider the type of work you will be doing on the laptop. If you are a gamer, you will need a high-performance laptop with a powerful graphics card. If you are a student, you may need a laptop with enough memory and storage to run multiple programs and store your files. Finally, you need to consider the size and weight of the laptop. Laptops can range from small and lightweight to large and bulky. Consider the portability of the laptop and how it will fit into your lifestyle. \\
\hline

P3A & Hey there! If you're in the market for a new laptop, I've got some advice for you. First things first, you need to decide what you're going to use it for. Are you a gamer? A professional graphic designer? A student? Knowing your needs will help you choose the right specs and features. Next, think about your budget. Laptops can range from cheap and cheerful to expensive and luxurious. Don't forget to factor in any accessories you might need, like a keyboard, mouse, or extra storage. And lastly, don't forget to consider the brand. Apple, Dell, HP, Lenovo - they're all good options, but they each have their own strengths and weaknesses. So, there you have it! My top tips for buying a new laptop. Don't forget to have fun with it, and don't be afraid to ask for help if you need it. Happy shopping! \\
\hline

P3B & If you're looking to buy a laptop, I suggest you do your research and compare prices from different retailers. Don't just settle for the first laptop you see, as there are many options available in the market. Consider your budget and the features you need, such as processing power, storage, and screen size. And don't forget to read reviews from other customers to get an idea of the product's quality. \\

\end{longtable}

\captionof{table}{Example 2: \textit{"Generate a recommendation for someone looking to buy a laptop."} with responses on different preference dimensions. } \label{tab:laptop}

\begin{longtable}{|p{0.12\textwidth}|p{0.83\textwidth}|}
\hline
\textbf{\textit{Instruction}} & \textit{In your words, what is meant by the economic concept of demand?} \\
\hline
\endfirsthead

\hline
\textbf{Preference Dimension} & \textbf{Output} \\
\hline
\endhead

\hline
\endfoot

\hline
\endlastfoot
\textbf{Preference Dimension} & \textbf{Output} \\

\hline

P1A & Demand is when people want to buy something and are willing to pay for it. It shows how much of a product or service people are willing to buy at a certain price. The more people want to buy something, the higher the demand for it. \\
\hline

P1B & Demand refers to the desire or willingness of consumers to purchase a particular good or service. It is a measure of the quantity of a product or service that consumers are willing and able to purchase at a given price. The concept of demand is central to economics, as it helps to determine the allocation of resources within an economy. The law of demand states that, all else being equal, as the price of a good or service increases, the quantity demanded will decrease, and as the price decreases, the quantity demanded will increase. This relationship between price and quantity demanded is a key factor in determining the optimal price for a product or service, as well as the overall level of production and consumption within an economy. \\
\hline

P2A & Demand refers to the desire and willingness of people to purchase a particular product or service, which is influenced by various factors such as price, quality, and availability. It is an essential concept in economics as it helps to determine the level of goods and services that will be produced and sold in an economy. \\
\hline

P2B & The economic concept of demand refers to the willingness and ability of individuals or organizations to purchase a specific good or service at a given price. Demand is influenced by various factors, including the price of the product, the income of the buyer, the availability of substitutes, and the level of consumer preferences. In economics, demand is represented by the demand curve, which shows the relationship between the price of a product and the quantity of that product that consumers are willing to purchase. The demand curve slopes downward, indicating that as the price of the product decreases, the quantity demanded will increase. Conversely, if the price of the product increases, the quantity demanded will decrease. The law of demand states that, all else being equal, an increase in price will result in a decrease in the quantity demanded, while an increase in income will lead to an increase in the quantity demanded. Similarly, a decrease in the price of a substitute product will lead to an increase in the quantity demanded of the original product, while an increase in the price of a complementary product will lead to a decrease in the quantity demanded of the original product. Demand is an important concept in economics as it helps to explain how markets function and how prices are determined. Understanding demand can help businesses and policymakers make informed decisions about pricing, production, and distribution of goods and services. \\
\hline

P3A & Demand, my friend, is like that one friend who's always there for you, no matter what. You know, the one who's always asking you to buy them stuff, even if you don't have the money. It's that friend who's always saying, "Hey, you gotta get me that new iPhone, it's so cool!" And even though you know you can't afford it, you still feel obligated to get it for them. That's demand, my friend. It's that insidious force that makes us spend money we don't have on things we don't need. It's like that one friend who's always trying to get you to go to their favorite restaurant, even though you've never liked their food. It's that friend who's always trying to get you to buy them the latest fashion trend, even though you know it's going to look ridiculous on them. That's demand, my friend. It's that insidious force that makes us spend money we don't have on things we don't need. \\
\hline

P3B & Demand refers to the desire and willingness of consumers to purchase a particular good or service. It is influenced by various factors such as price, income, tastes and preferences, and the availability of substitute goods. The higher the demand for a product, the more likely it is for firms to produce and sell that product, and for consumers to purchase it. The concept of demand is important in economics as it helps to determine the quantity of a good or service that will be produced and sold in an economy. \\

\end{longtable}

\captionof{table}{Example 3: \textit{"In your words, what is meant by the economic concept of demand?"} with responses on different preference dimensions. } \label{tab:demand}

\end{document}